\def\paperTitle{Attention Transfer Is Not Universally Effective for Vision Transformers}

\def\authorBlock{
{\bfseries\fontsize{11.2}{16}\selectfont
    Huaiyuan Qin\,\textnormal{\textsuperscript{1}}%
\hspace{12pt}
    Muli Yang\,\textnormal{\textsuperscript{1}}%
\hspace{12pt}
    Gabriel James Goenawan\,\textnormal{\textsuperscript{2,$\dagger$}}%
}\vspace{3pt}\\

{\bfseries\fontsize{11.2}{16}\selectfont
    Peng Hu\,\textnormal{\textsuperscript{3}}%
\hspace{12pt}
    Chen Gong\,\textnormal{\textsuperscript{4}}%
\hspace{12pt}
    Xi Peng\,\textnormal{\textsuperscript{3}}%
\hspace{12pt}
    Hongyuan Zhu\,\textnormal{\textsuperscript{1,\Letter}}%
}
\vspace{4pt}
\\ 

{\fontsize{10.9}{16}\selectfont
\textsuperscript{1}Institute of Advanced Intelligence and Computing (IAIC), A*STAR, Singapore
}
\\

{\fontsize{10.9}{16}\selectfont
\textsuperscript{2}Intrinsic, Google
}\hspace{6pt}
{\fontsize{10.9}{16}\selectfont
\textsuperscript{3}School of Artificial Intelligence, Sichuan University
}
\\

{\fontsize{10.9}{16}\selectfont
\textsuperscript{4}School of Automation and Intelligent Sensing, Shanghai Jiao Tong University
}\vspace{3pt}\\

{\tt
\small
\{qinhy, yangml, zhuh\}@a-star.edu.sg,
gabrieljames@google.com,
\hspace{-6.28pt}}\\

{\tt
\small
\{penghu.ml, pengx.gm\}@gmail.com,
chen.gong@sjtu.edu.cn
\hspace{-6.28pt}}\vspace{6pt}\\

\href{https://leonqin430.github.io/projects/attention-transfer-universally/}{\textbf{\page~~Project Page}}
\vspace{-6pt}

}

\newif\ifreview 
\newif\ifarxiv 
\newif\ifcamera \newcommand{\cameraready}{\cameratrue}

\cameraready

\pdfoutput=1
\documentclass{article}

\PassOptionsToPackage{numbers, compress}{natbib}
 
\ifreview \usepackage{meta/neurips_2026} \fi
\ifarxiv \usepackage[preprint]{meta/neurips_2026} \fi
\ifcamera \usepackage[main, final]{meta/neurips_2026} \fi

\usepackage{fix-cm}
\usepackage{array}
\usepackage{bbm}
\usepackage{nicematrix}

\usepackage{tipa}
\usepackage{dsfont}
\usepackage{etoolbox}  

\usepackage[noend]{algorithmic}
\usepackage{algorithm}

\usepackage{float}
\usepackage{newfloat}
\usepackage{listings}
\floatstyle{ruled}
\newfloat{listing}{tb}{lst}{}
\floatname{listing}{\small Algorithm}

\definecolor{mygray}{RGB}{234,234,234}

\newcommand{\ra}[1]{\renewcommand{\arraystretch}{#1}}

\usepackage{adjustbox}
\usepackage[table,dvipsnames]{xcolor}

\definecolor{darkgreen}{rgb}{0.13, 0.55, 0.13}

\usepackage{graphicx}	
\usepackage{amsmath}	
\usepackage{amssymb}	
\usepackage{booktabs}
\usepackage{times}
\usepackage{epsfig}
\usepackage{caption}
\usepackage{float}
\usepackage{placeins}
\usepackage{color, colortbl}
\usepackage{stfloats}
\usepackage{enumitem}
\usepackage{tabularx}
\usepackage{xstring}
\usepackage{multirow}
\usepackage{xspace}
\usepackage{url}
\usepackage{subcaption}
\usepackage{xcolor}

\usepackage{inconsolata}

\ifcamera \usepackage[accsupp]{axessibility} \fi

% ...

\ifarxiv  \fi

\newcommand{\R}[1]{{%
    \textbf{%
        \ifstrequal{#1}{1}{\textcolor{red}{R#1}}{%
        \ifstrequal{#1}{2}{\textcolor{blue}{R#1}}{%
        \ifstrequal{#1}{3}{\textcolor{magenta}{R#1}}{%
        \ifstrequal{#1}{4}{\textcolor{teal}{R#1}}{%
                           \textcolor{cyan}{R#1}%
        }}}}%
    }%
}}

\definecolor{Gray}{gray}{0.5}
\definecolor{nicergreen}{rgb}{0.13, 0.54, 0.13}
\definecolor{nicered}{rgb}{0.83, 0.16, 0.16}
\definecolor{lightgray}{RGB}{230, 230, 230}
\definecolor{Highlight}{HTML}{39b54a}  %

\usepackage{appendix}
\usepackage{footnote}

\usepackage{microtype}
\usepackage{cuted}
\usepackage{tocloft}

\usepackage[T1]{fontenc}
\usepackage{DejaVuSans}

\usepackage{ifsym, marvosym}

\let\svthefootnote\thefootnote
\newcommand\freefootnote[1]{%
  \let\thefootnote\relax%
  \footnotetext{#1}%
  \let\thefootnote\svthefootnote%
}

\makeatletter
\DeclareRobustCommand\onedot{\futurelet\@let@token\@onedot}
\def\@onedot{\ifx\@let@token.\else.\null\fi\xspace}

 \def\vs{\emph{vs}\onedot}

\makeatother

\usepackage{wrapfig}  

\usepackage{xr-hyper}

\makeatletter
\newcommand*{\addFileDependency}[1]{
  \typeout{(#1)}
  \@addtofilelist{#1}
  \IfFileExists{#1}{}{\typeout{No file #1.}}
}

\makeatother

\definecolor{cvprblue}{rgb}{0.21,0.49,0.74}
\usepackage[pagebackref,breaklinks,colorlinks]{hyperref}
\usepackage[capitalize]{cleveref}
\crefname{section}{Sec.}{Secs.}
\crefname{table}{Table}{Tables}
\crefname{figure}{Fig.}{Figs.}

\usepackage[utf8]{inputenc} 
\usepackage[T1]{fontenc}    
\usepackage{url}            
\usepackage{booktabs}       
\usepackage{amsfonts}       
\usepackage{amssymb}        
\usepackage{nicefrac}       
\usepackage{microtype}      
\usepackage{xcolor}         

\def\page{\raisebox{-1.5pt}{\includegraphics[height=1.05em]{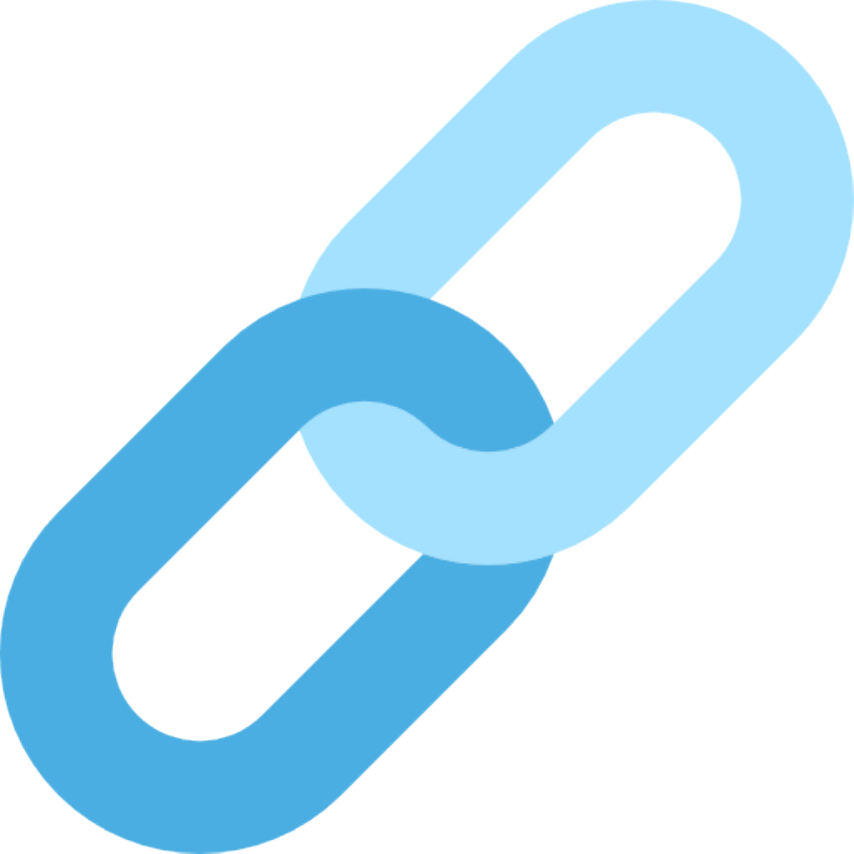}}}

\begin{document}
\title{\paperTitle}
\author{\authorBlock}
\maketitle

\freefootnote{\hspace{-12pt}\textsuperscript{$\dagger$} This paper was completely accomplished when GJG was at A*STAR.}
\freefootnote{\hspace{-12pt}\textsuperscript{\Letter} Corresponding author.}

\begin{abstract}
A recent work shows that Attention Transfer, which transfers only the attention patterns from a pre-trained teacher Vision Transformer (ViT) to a randomly initialized standard student ViT, is sufficient to recover the full benefit of the teacher's pre-trained weights.
We revisit this finding on a comprehensive benchmark of 20 teachers from 11 well-known ViT families and reveal that Attention Transfer is not universally effective. 
While 7 families transfer successfully, 4 consistently fail, falling up to 5.1\% below the from-scratch no-transfer baseline. 
Further results demonstrate that this failure is family-consistent across model sizes, and persists under extended training durations, different transfer datasets, and out-of-distribution evaluations.
Controlled analyses then consistently localize the problem to the attention-routing channel, indicating that the key issue is not whether the student can match the teacher's attention patterns, but whether the matched patterns remain functional for the student.
Crucially, we identify architectural mismatch between the pre-trained teacher and the standard student as the primary mechanism.
By adding only the teacher's native architectural components to the student in a randomly initialized state, we completely reverse the failure for all 4 families. 
Notably, these components alone do not improve from-scratch training, confirming that they specifically unlock the usability of the teacher's attention.
We further show that this failure is not explained by the inadequate choice of transfer loss, differences in pre-training recipes, or the intrinsic quality of attention maps.
Our findings refine the prevailing understanding of attention in ViT representations: attention is sufficient \textit{only} when the student architecture matches the teacher.

\end{abstract}

\vspace{-6pt}
\section{Introduction}
\label{sec:intro}

Pre-trained Vision Transformers (ViTs)~\cite{dosovitskiy2020image} have become the standard foundation for modern visual representation learning, but obtaining such weights typically requires massive computation and large curated datasets~\cite{oquab2023dinov2,simeoni2025dinov3,radford2021learning}.
This makes the efficient reuse of pre-trained representations a central objective in computer vision.
In practice, transferring a pre-trained ViT usually means initializing a downstream model with the full set of teacher weights.
While effective, it remains fundamentally unclear which specific components of a pre-trained ViT actually encode its transferable knowledge~\cite{walmer2023teaching,park2023what,dravid2023rosetta}.

\begin{figure}[t]
    \centering
    \includegraphics[width=0.98\linewidth]{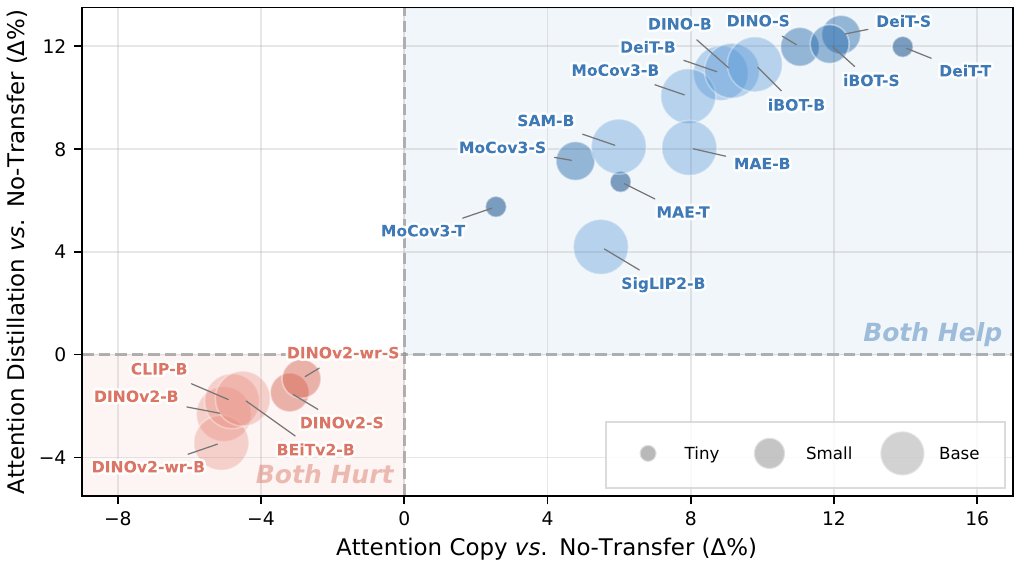}
    \vspace{-0.7em}
    \caption{\textbf{Attention Transfer is not universally effective.}
    We begin by evaluating Attention Transfer~\cite{li2024attention} on the broader model zoo of 20 pre-trained teachers from 11 well-known ViT families under the default ImageNet-1K setup.
    Each point represents one set of pre-trained weights; the axes report $\Delta$ Top-1 accuracy (\%) relative to the No-Transfer baseline under two transfer methods: \textit{Attention Copy} (x-axis) and \textit{Attention Distillation} (y-axis).
    Marker size denotes the model size (Tiny/Small/Base).
    7 families consistently succeed (in the {\color[HTML]{2166ac}blue} top-right region), while 4 consistently fail (in the {\color[HTML]{d6604d}red} bottom-left region) under \textit{both} transfer methods.
    }
    \vspace{-1.5em}
    \label{fig:main_results}
\end{figure}

One recent work~\cite{li2024attention} offers a striking answer to this question.
It shows that \emph{Attention Transfer}, which transfers only the teacher's attention patterns to a randomly initialized standard student ViT through attention copy or distillation, is sufficient to recover the full benefit of the teacher's pre-trained weights.
This result suggests that the essential transferable structure of a pre-trained ViT lies in its inter-token routing patterns, while the remaining parameters can be learned from scratch once this attention behavior is imposed. 
A natural question is whether this finding holds broadly across the increasingly diverse landscape of pre-trained ViTs, and more fundamentally, whether attention patterns are truly a portable representation of pre-trained knowledge across architectures.

In this work, we show that this \textit{\textbf{does not}} hold universally.
By revisiting Attention Transfer on a substantially broader and more modern model zoo of 20 teachers from 11 well-known ViT families~\cite{oquab2023dinov2,radford2021learning,touvron2021training,caron2021emerging,chen2021empirical,he2022masked,zhou2021ibot,peng2022beit,tschannen2025siglip,kirillov2023segment,darcet2023vision}, we uncover a hidden validity boundary that was invisible under the evaluations in~\cite{li2024attention}: its effectiveness is highly family-dependent.
Under the standard teacher-student transfer protocol used in~\cite{li2024attention}, Attention Transfer consistently succeeds for 7 families, but consistently fails for 4.
For these failure families, transferring attention yields strictly worse downstream performance than the from-scratch no-transfer baseline, with the worst case falling 5.1\% below scratch.
Further results show that this failure is family-consistent across model sizes, and persists under extended training durations, different transfer datasets, and out-of-distribution evaluations.

To understand why these failures arise, we conduct a set of controlled analyses, including component-level, layer-wise, and QKV decomposition ablations, which consistently localize the problem to the attention-routing channel, the part that controls inter-token information flow in ViTs, under a standard student architecture.
This shows that the key issue is not simply whether the student can match the teacher's attention patterns, but whether the matched patterns remain functional in the student's attention routing.
This naturally raises a sharper question: what determines this functional usability across teacher families?

Crucially, we identify \textit{architectural mismatch} between the pre-trained teacher and the standard student as the missing factor.
The standard student ViT used in~\cite{li2024attention} omits architectural components that are native to those failure families, such as \texttt{LayerScale} in DINOv2~\cite{oquab2023dinov2} and DINOv2-wr\footnote{We term the family of DINOv2-with-registers~\cite{darcet2023vision} as DINOv2-wr in the following for simplicity.}~\cite{darcet2023vision}, \texttt{PreLayerNorm} in CLIP~\cite{radford2021learning}, and \texttt{LayerScale} together with \texttt{RelativePositionBias} in BEiTv2~\cite{peng2022beit}.
Moreover, by simply adding the teacher's native architectural components to the student in a \textit{randomly initialized} state, we completely reverse this failure for all 4 families. 
Importantly, training this native-architecture student entirely from scratch yields worse performance than a standard from-scratch student, confirming that these components do not artificially boost performance on their own.
Instead, they specifically make the teacher's transferred attention usable, showing that attention is sufficient to recover the full benefit of the teacher's pre-trained weights \textit{only} when the student architecture matches the teacher.

We then ask whether simpler explanations could account for these failures.
We investigate three natural alternative explanations: the choice of inadequate transfer loss, differences in pre-training recipes, and the intrinsic quality of attention maps.
First, through empirical analysis, we find that different losses with sufficient magnitude can induce similar supervision signals for Attention Transfer.
However, for the failure families, increasing the magnitude of any loss further hurts performance, showing that the loss itself does not explain the failure.
Second, we analyze several hypotheses based on common properties in pre-training recipes of the failure families, including different training signals, different training data sources, and other known special attributes, but find none that explain this family-level boundary.
Lastly, we measure the quality of the teachers' attention maps by evaluating their object localization accuracy, finding that the quality ranking does not predict the observed failure, and that the native-architecture rescue, which reverses the failure, does not consistently change this quality.

In summary, our primary contributions are as follows:
\begin{enumerate}
    \item \textbf{Failure Discovery.} We provide the first comprehensive Attention Transfer~\cite{li2024attention} benchmark across 20 teachers and 11 well-known ViT families, uncovering a critical validity boundary where 4 major families consistently fail to transfer to a standard student architecture.
    \item \textbf{Three-Level Localization.} Through systematic component-level, layer-wise, and QKV decomposition ablations, we localize the failure specifically to the attention-routing channel, showing that the key issue is whether the matched patterns remain functional for the student.
    \item \textbf{Architectural Compatibility as Mechanism \& Fix.} We identify architectural mismatch as the primary mechanism behind Attention Transfer failure.
    Adding the teacher's native components to the student in a randomly initialized state completely reverses the failure for all 4 families, while these components alone hurt from-scratch training.
    \item \textbf{Elimination of Alternative Explanations.} We analytically and empirically rule out potential inadequate choice of transfer loss, pre-training recipe differences, and the intrinsic quality of attention maps as explanations for the observed failure boundary, strengthening the finding that architectural mismatch is the primary explanation of Attention Transfer failure.
\end{enumerate}

\vspace{-6pt}
\section{Preliminaries}
\label{sec:preliminary}

\vspace{-3pt}
\subsection{Attention Transfer}
\vspace{-3pt}
Following~\cite{li2024attention}, we study \emph{Attention Transfer}, where a randomly initialized student Vision Transformer (ViT)~\cite{dosovitskiy2020image} is trained to inherit the full benefit of a pre-trained teacher without loading the teacher's full pre-trained weights.
Specifically, for a ViT block $\ell$, given $Q$, $K$, and $V$ as the corresponding queries, keys, and values, the attention function is defined as\footnote{We omit the scaling factor $1/\sqrt{d_k}$ for simplicity.}:
\begin{align}
\label{eq:attn}
    f_\text{attn}^{(\ell)} =\underbrace{\mathrm{softmax}\left(Q^{(\ell)}K^{(\ell)\top}\right)}_{\text{attention map}}V^{(\ell)},
\end{align}
where the softmax function is computed per query for the \emph{attention map}.

We study two forms of Attention Transfer from a pre-trained teacher ViT to a standard student\footnote{The standard student denotes the vanilla ViT architecture of~\cite{dosovitskiy2020image,touvron2021training}, which is kept fixed regardless of the teacher's architecture; only the teacher's attention is copied or distilled into it.} of the same size:
\textbf{\textit{Attention Copy}}, which directly copies the teacher's attention projection weights and keeps them frozen; and \textbf{\textit{Attention Distillation}}, which trains the student to match the teacher's attention patterns through an auxiliary transfer objective. 
Given $Q_s{K_s^\top}$ and $Q_t{K_t^\top}$ from the student and teacher, respectively, the transfer loss at block $\ell$ is defined as: 
\begin{align}
    \mathcal{L}_\text{transfer}^{(\ell)} = \mathcal{L}\left[\mathrm{softmax}\left(Q_s^{(\ell)}{K_s^{(\ell)\top}}\right),\, \mathrm{softmax}\left(Q_t^{(\ell)}{K_t^{(\ell)\top}}\right)\right],
\end{align}
where $\mathcal{L}$ denotes Cross-Entropy loss by default in~\cite{li2024attention}.
%

\subsection{Experimental Setup}
\label{subsec:exp_setup}

\paragraph{Model Zoo.}
We revisit Attention Transfer on a comprehensive model zoo which consists of 20 pre-trained weights from 11 well-known ViT families, including DeiT~\cite{touvron2021training}, DINO~\cite{caron2021emerging}, MoCov3~\cite{chen2021empirical}, MAE~\cite{he2022masked}, iBOT~\cite{zhou2021ibot}, SigLIP2~\cite{tschannen2025siglip}, SAM~\cite{kirillov2023segment}, DINOv2~\cite{oquab2023dinov2}, DINOv2-wr~\cite{darcet2023vision}, CLIP~\cite{radford2021learning}, and BEiTv2~\cite{peng2022beit}, spanning three model sizes: Tiny, Small, and Base.

\vspace{-6pt}
\paragraph{Evaluation Protocol.}
Unless specifically mentioned, all Attention Transfer experiments are conducted on ImageNet-1K classification~\cite{deng2009imagenet} with the default protocol in~\cite{li2024attention} using the standard teacher-student architecture.
To carefully isolate the effects of the transferred attention patterns, we maintain a strictly controlled training protocol across all experiments: models are trained for 20 epochs with the AdamW optimizer~\cite{loshchilov2017decoupled}.
Other training hyper-parameters follow the defaults in~\cite{li2024attention}.
All reported results are averaged over 3 random seeds to ensure statistical significance.
More details are in~\Cref{sec:appendix_implementation_details}.

\vspace{-6pt}
\paragraph{No-Transfer Baseline.}
As a primary reference, we define the \emph{No-Transfer} (NoT) baseline, where the same standard ViT is randomly initialized and trained from scratch for 20 epochs.
Attention Transfer from one given pre-trained weights is considered a ``failure'' when the final accuracy falls below this baseline, indicating that imposing attention patterns from the teacher is less effective than learning from scratch under the same student architecture.

\section{Attention Transfer Is Not Universally Effective}
\label{sec:main_results}

\subsection{Main Results}
We begin by evaluating the two Attention Transfer methods on the full benchmark of 20 pre-trained teachers from 11 well-known ViT families under the default ImageNet-1K setup.
\Cref{fig:main_results} summarizes the performance of both \emph{Attention Copy} and \emph{Attention Distillation}, measured as Top-1 accuracy gap relative to the No-Transfer baseline.
Points in the upper-right quadrant therefore indicate teachers for which both transfer methods are beneficial, whereas points in the lower-left (failure) quadrant indicate teachers for which both transfer methods are harmful.

The results reveal a clear family-dependent split, and this split boundary is consistent across model sizes rather than driven by isolated pre-trained weights.
All pre-trained weights maintain a consistent sign under both Attention Transfer methods across all seeds.
Specifically, 7 families, including DeiT, DINO, MoCov3, MAE, iBOT, SigLIP2, and SAM, improve over the No-Transfer baseline, with gains reaching 13.9\% for DeiT-T under Attention Copy.
In contrast, 4 families: DINOv2, DINOv2-wr, CLIP, and BEiTv2, consistently fall into the failure region.
The worst case is DINOv2-wr-B, which underperforms the No-Transfer baseline by 5.1\% under Attention Copy, showing that imposing the attention patterns of certain teachers can substantially degrade learning under the same student architecture.
Notably, all teachers evaluated in~\cite{li2024attention} belong exclusively to our success group, while the failure families are absent from their evaluations.
These results show that Attention Transfer is not universally effective under the standard teacher-student transfer protocol.

\begin{wrapfigure}{r}{0.48\textwidth}
    \vspace{-2.2em}
    \includegraphics[width=0.98\linewidth]{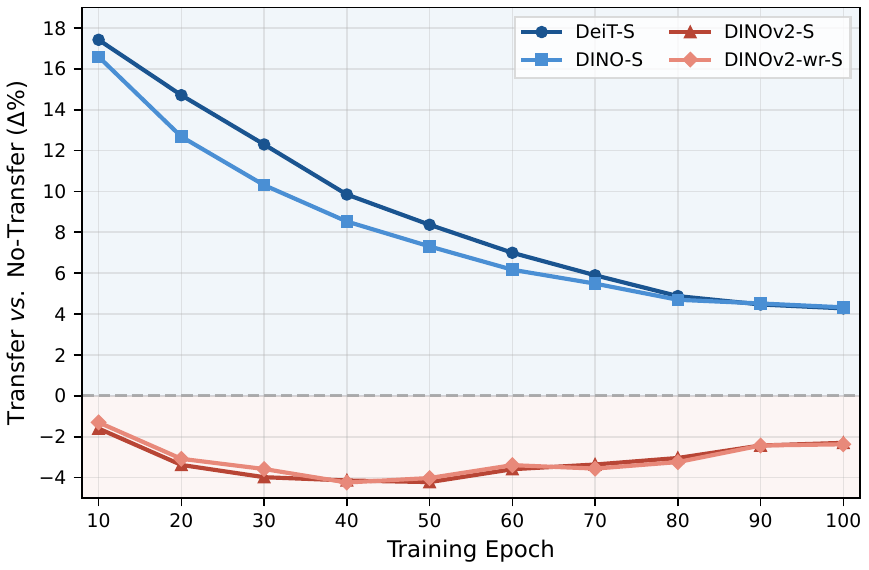}
    \vspace{-0.5em}
    \caption{\textbf{Extended training robustness.}
    We extend the Attention Distillation training up to 100 epochs and report the gap relative to No-Transfer every 10 epochs.
    The same success/failure split remains consistent with the results in~\Cref{fig:main_results}.
    }
    \label{fig:100ep_gap}
    \vspace{-2.3em}
\end{wrapfigure}

\subsection{Robustness of the Failure}
A natural counter-hypothesis is that the failure families might suffer from slower convergence under the transferred attention patterns, thus extending the training duration or changing the transfer dataset could eventually close or even reverse the performance gap. 
To rule this out, we further rigorously test the robustness of this observed failure using Attention Distillation on a representative subset from two success (DeiT, DINO) and two failure (DINOv2, DINOv2-wr) families with model size Small.

\vspace{-6pt}
\paragraph{Extended Training Robustness.}
We extend the training schedule from 20 to 100 epochs for both the success and failure families. 
As shown in~\Cref{fig:100ep_gap}, the performance gap persists throughout the entire extended learning process. 
While successful families maintain strong positive gains over the No-Transfer baseline, the failure families remain consistently negative even with substantially longer training.
This confirms that the observed success/failure split is robust to extended training and that the 20-epoch results in~\Cref{fig:main_results} are sufficient to capture this boundary.

\vspace{-6pt}
\paragraph{Dataset Robustness.}
We then test whether the observed failure is merely an artifact of the ImageNet-1K transfer dataset.
\Cref{tab:inat} reports the performance of Attention Distillation on three fine-grained iNaturalist variants~\cite{van2018inaturalist} under the standard 20-epoch protocol. 
The same success/failure split appears consistently across all three year splits, matching the ImageNet-1K results. 
This shows that the failure of certain families is robust across different datasets.

\vspace{-6pt}
\paragraph{Out-of-Distribution Robustness.}
This observed success/failure split also persists under the out-of-distribution evaluation.
\Cref{tab:ood} shows that on four ImageNet-1K distribution-shift benchmarks~\cite{hendrycks2021many,hendrycks2021natural,wang2019learning,recht2019imagenet}, Attention Distillation preserves the same positive/negative pattern observed in the in-distribution settings.  
This further confirms that the family-consistent failure of Attention Transfer is robust to substantial distribution shift.

\begin{table}[t]
    \begin{minipage}[t]{0.48\textwidth}
        \centering
\small
\caption{\textbf{Dataset robustness.}
We evaluate Attention Distillation on iNaturalist~\cite{van2018inaturalist} with 20-epoch training.
\textcolor{gray}{Gray row} shows the default ImageNet-1K results for reference.
Colored deltas indicate gains/drops relative to No-Transfer (NoT) baseline.
The success/failure split remains consistent with different datasets.
}
\vspace{3pt}
\label{tab:inat}
\ra{1.02}
\setlength{\tabcolsep}{1.6pt}
\footnotesize
\begin{tabular}{l c cccc}
 & \cellcolor[HTML]{d8d8d8}NoT & \color[HTML]{2166ac}DeiT & \color[HTML]{2166ac}DINO & \color[HTML]{d6604d}DINOv2 & \color[HTML]{d6604d}DINOv2-wr \\
\midrule
\color{gray} IN-1K & \cellcolor[HTML]{d8d8d8}\color{gray}63.1 & \color{gray} 75.6 {\scriptsize\color[HTML]{2166ac}+12.4} & \color{gray} 75.1 {\scriptsize\color[HTML]{2166ac}+12.0} & \color{gray} 61.7 {\scriptsize\color[HTML]{d6604d}-1.5} & \color{gray} 62.2 {\scriptsize\color[HTML]{d6604d}-0.9} \\
iNat 2017 & \cellcolor[HTML]{d8d8d8}26.6 & 44.7 {\scriptsize\color[HTML]{2166ac}+18.1} & 46.0 {\scriptsize\color[HTML]{2166ac}+19.4} & 24.4 {\scriptsize\color[HTML]{d6604d}-2.2} & 24.5 {\scriptsize\color[HTML]{d6604d}-2.1} \\
iNat 2018 & \cellcolor[HTML]{d8d8d8}35.0 & 50.2 {\scriptsize\color[HTML]{2166ac}+15.2} & 51.2 {\scriptsize\color[HTML]{2166ac}+16.3} & 33.1 {\scriptsize\color[HTML]{d6604d}-1.9} & 33.8 {\scriptsize\color[HTML]{d6604d}-1.2} \\
iNat 2019 & \cellcolor[HTML]{d8d8d8}40.1 & 56.8 {\scriptsize\color[HTML]{2166ac}+16.7} & 55.7 {\scriptsize\color[HTML]{2166ac}+15.5} & 38.1 {\scriptsize\color[HTML]{d6604d}-2.1} & 38.7 {\scriptsize\color[HTML]{d6604d}-1.4} \\
\end{tabular}

    \end{minipage}
    \hfill
    \begin{minipage}[t]{0.48\textwidth}
        \centering
\small
\caption{\textbf{Out-of-distribution robustness.}
We evaluate Attention Distillation on four distribution-shift benchmarks~\cite{hendrycks2021many,hendrycks2021natural,wang2019learning,recht2019imagenet} with 100-epoch training.
Colored deltas indicate gains/drops relative to No-Transfer (NoT) baseline.
The success/failure split remains consistent with the in-distribution evaluation.
}
\vspace{3pt}
\label{tab:ood}
\ra{1.02}
\setlength{\tabcolsep}{1.8pt}
\footnotesize
\begin{tabular}{l c cccc}
 & \cellcolor[HTML]{d8d8d8}NoT & \color[HTML]{2166ac}DeiT & \color[HTML]{2166ac}DINO & \color[HTML]{d6604d}DINOv2 & \color[HTML]{d6604d}DINOv2-wr \\
\midrule
IN-A~\cite{hendrycks2021natural}     & \cellcolor[HTML]{d8d8d8}9.2  & 17.6 {\scriptsize\color[HTML]{2166ac}+8.4}  & 17.2 {\scriptsize\color[HTML]{2166ac}+8.0}  & 6.0 {\scriptsize\color[HTML]{d6604d}-3.2}  & 5.8 {\scriptsize\color[HTML]{d6604d}-3.4} \\
IN-R~\cite{hendrycks2021many}     & \cellcolor[HTML]{d8d8d8}35.9 & 42.3 {\scriptsize\color[HTML]{2166ac}+6.4}  & 42.8 {\scriptsize\color[HTML]{2166ac}+6.9}  & 32.8 {\scriptsize\color[HTML]{d6604d}-3.1} & 32.2 {\scriptsize\color[HTML]{d6604d}-3.7} \\
IN-S~\cite{wang2019learning}     & \cellcolor[HTML]{d8d8d8}22.1 & 29.7 {\scriptsize\color[HTML]{2166ac}+7.6}  & 29.4 {\scriptsize\color[HTML]{2166ac}+7.3}  & 19.5 {\scriptsize\color[HTML]{d6604d}-2.6} & 19.0 {\scriptsize\color[HTML]{d6604d}-3.1} \\
IN-V2~\cite{recht2019imagenet}    & \cellcolor[HTML]{d8d8d8}63.9 & 69.5 {\scriptsize\color[HTML]{2166ac}+5.5}  & 69.9 {\scriptsize\color[HTML]{2166ac}+6.0}  & 61.4 {\scriptsize\color[HTML]{d6604d}-2.5} & 61.6 {\scriptsize\color[HTML]{d6604d}-2.4} \\
\end{tabular}

    \end{minipage}
    \vspace{-1.2em}
\end{table}

\section{Localizing the Failure}
\label{sec:localize}

\vspace{-3pt}
Having established the robust failure of Attention Transfer for specific teacher families, we now investigate exactly \textit{where} the transferred pre-trained weights harm the student.
We diagnose this problem systematically through the three-level analysis below.

\begin{wrapfigure}[23]{r}{0.48\textwidth}
    \vspace{-1.2em}
    \includegraphics[width=0.98\linewidth]{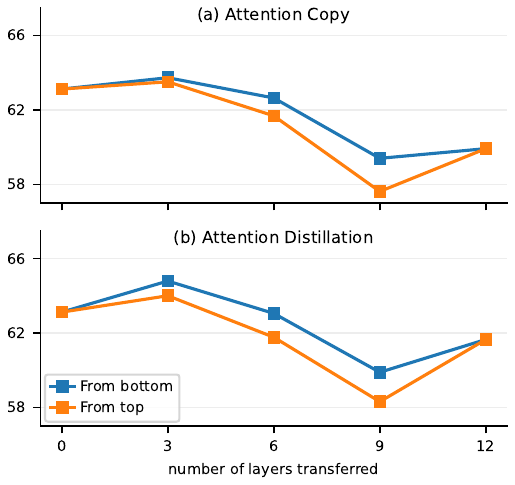}
    \vspace{-0.3em}
    \caption{\textbf{Layer-wise decomposition.}
    We transfer attention from the top or bottom $k$ layers of DINOv2-S under both Attention Copy and Attention Distillation.
    In contrast to the monotonic gains in~\cite{li2024attention}, no layer subset reproduces success-family gains, and harmful transfer appears across every transfer depth for this failure teacher.
    }
    \label{fig:layerwise}
    \vspace{-1.0em}
\end{wrapfigure}

\vspace{-6pt}
\paragraph{Component-Level Decomposition.}

We first ask whether the family-specific split is driven by a particular ViT component or reflects a generic difference in the usefulness of the pre-trained weights.
To test this, we selectively initialize the attention and MLP blocks from representative teachers under a component-level decomposition setup with full fine-tuning.
As shown in~\Cref{tab:component}, MLP-only initialization is consistently beneficial for both the success teacher (DeiT-S) and the failure teachers (DINOv2-S and CLIP-B), aligning with the findings in~\cite{geva2021transformer}.
In contrast, Attention-only initialization preserves the success/failure split. 
This shows that the failure families do not suffer from a generic inability to benefit from the pre-trained weights as a whole.
Instead, the failure is consistently localized to the attention pathway, motivating a finer-grained analysis of where this harm actually appears within attention itself.

\vspace{-6pt}
\paragraph{Layer-wise Decomposition.}

We therefore ask whether the harmful effect of Attention Transfer is concentrated at a particular depth.
To test this, we transfer attention from the top or bottom $k$ layers of one representative failure teacher, DINOv2-S, under both Attention Copy and Attention Distillation.
As shown in~\Cref{fig:layerwise}, no layer subset recovers the positive gains observed for success families.
Instead, performance remains below the No-Transfer baseline across all tested transfer depths, in clear contrast to the observed monotonic-gain pattern for success teachers in~\cite{li2024attention}.
This rules out a single problematic depth as the source of failure and suggests that the failure is distributed broadly across the transferred attention patterns.

\vspace{-6pt}
\paragraph{QKV Decomposition.}

We finally ask whether the failure can be attributed to a single attention subcomponent.
Across both Attention Copy and Attention Distillation,~\Cref{tab:qkv} shows that transferring any of $Q$, $K$, or $V$ alone remains beneficial for the success teacher DeiT-S, consistent with the findings in~\cite{li2024attention}.
However, the same single-component transfer is harmful for the failure teacher DINOv2-S, with $V$ producing the largest drop.
Together with the previous layer-wise analysis, this further confirms that the failure is not confined to one specific layer or one isolated attention subcomponent, but is distributed broadly throughout the overall attention pathway.

This directly motivates the next question: why does transferred attention remain functional for some teacher families, yet become ineffective for others under the same standard student architecture?

\begin{table}[t]
    \begin{minipage}[t]{0.48\textwidth}
        \centering
\small
\caption{\textbf{Component-level decomposition.}
We selectively initialize attention (Attn.) and MLP blocks from three representative teachers and fine-tune end-to-end.
\checkmark\ denotes initialized components, and blank cells denote random initialization.
Attention-only preserves success/failure split, while MLP-only is uniformly strong across all three families.
}
\vspace{6pt}
\label{tab:component}
\ra{1.02}
\setlength{\tabcolsep}{4pt}
\footnotesize
\begin{tabular}{cc | ccc}
Attn. & MLP & \color[HTML]{2166ac}DeiT-S & \color[HTML]{d6604d}DINOv2-S & \color[HTML]{d6604d}CLIP-B \\
\midrule
 &
  & 66.0
  & 65.3
  & 70.3 \\
\checkmark &
  & 71.7 {\scriptsize\color[HTML]{2166ac}+5.7}
  & 63.7 {\scriptsize\color[HTML]{d6604d}-1.6}
  & 69.4 {\scriptsize\color[HTML]{d6604d}-0.9} \\
 & \checkmark
  & 77.7 {\scriptsize\color[HTML]{2166ac}+11.7}
  & 69.2 {\scriptsize\color[HTML]{2166ac}+3.9}
  & 74.7 {\scriptsize\color[HTML]{2166ac}+4.4} \\
\checkmark & \checkmark
  & 79.3 {\scriptsize\color[HTML]{2166ac}+13.3}
  & 71.0 {\scriptsize\color[HTML]{2166ac}+5.7}
  & 80.1 {\scriptsize\color[HTML]{2166ac}+9.8} \\
\end{tabular}

    \end{minipage}
    \hfill
    \begin{minipage}[t]{0.48\textwidth}
       \centering
\small
\caption{\textbf{QKV decomposition.}
We replace the default full-attention transfer with subset transfer over $Q$, $K$, and $V$ under both methods.
All alternatives improve performance on the success family but remain negative and even worse on the failure family, indicating the failure does not lie in any specific attention subcomponent.
}
\vspace{3pt}
\label{tab:qkv}
\ra{1.02}
\setlength{\tabcolsep}{4pt}
\footnotesize
\begin{tabular}{c cc cc}
 & \multicolumn{2}{c}{\color[HTML]{2166ac}{DeiT-S} \color{gray}{(NoT: 63.1)}}  & \multicolumn{2}{c}{\color[HTML]{d6604d}DINOv2-S \color{gray}{(NoT: 63.1)}} \\
\cmidrule(lr){2-3} \cmidrule(lr){4-5}
 & Copy & Distill & Copy & Distill \\
\midrule
$Q$    & 78.1 {\scriptsize\color[HTML]{2166ac}+15.0} & 77.0 {\scriptsize\color[HTML]{2166ac}+13.9} & 57.9 {\scriptsize\color[HTML]{d6604d}-5.2}  & 59.0 {\scriptsize\color[HTML]{d6604d}-4.1} \\
$K$    & 78.7 {\scriptsize\color[HTML]{2166ac}+15.6} & 76.7 {\scriptsize\color[HTML]{2166ac}+13.6} & 57.4 {\scriptsize\color[HTML]{d6604d}-5.7}  & 58.7 {\scriptsize\color[HTML]{d6604d}-4.4} \\
$V$    & 78.4 {\scriptsize\color[HTML]{2166ac}+15.3} & 78.0 {\scriptsize\color[HTML]{2166ac}+14.9} & 47.3 {\scriptsize\color[HTML]{d6604d}-15.9} & 54.9 {\scriptsize\color[HTML]{d6604d}-8.3} \\
Attn.   & 75.3 {\scriptsize\color[HTML]{2166ac}+12.2} & 75.6 {\scriptsize\color[HTML]{2166ac}+12.4} & 59.9 {\scriptsize\color[HTML]{d6604d}-3.2}  & 61.7 {\scriptsize\color[HTML]{d6604d}-1.5} \\
\end{tabular}

    \end{minipage}
    \vspace{-1.4em}
\end{table}

\section{Architectural Compatibility Is the Key}
\label{sec:solution}

\vspace{-3pt}
Having localized the failure to the attention pathway, one question remains: what determines whether transferred attention is functional across teacher families?
We address this below, starting with a simple native-architecture rescue.

\subsection{Native-architecture Rescue}

\paragraph{Motivation.}
Results in~\cite{li2024attention} suggest that transferring \textit{only} the attention can recover the full benefit of the teacher's pre-trained weights under a standard teacher-student architecture.
However, we revisit this claim in~\Cref{sec:main_results} to find that this does not hold universally, and systematically localize the failure to the attention pathway in~\Cref{sec:localize}, further confirming that this failure is broadly distributed across attention layers and Q/K/V terms.
This leads us to hypothesize that the missing factor is \emph{architectural compatibility} in the default protocol: transferred attention remains usable only when the student preserves the architectural components under which the teacher originally learned its attention patterns.
Concretely, each failure family in~\Cref{fig:main_results} contains one or more such components that are absent from the standard student architecture: \texttt{LayerScale}\footnote{\texttt{LayerScale} was originally introduced in CaiT~\cite{touvron2021going} to stabilize the training of very deep ViTs, and is not adopted by the ViT implementations of~\cite{dosovitskiy2020image,touvron2021training}, which is used in the standard teacher-student architecture.} in DINOv2~\cite{oquab2023dinov2} and DINOv2-wr~\cite{darcet2023vision}, \texttt{PreLayerNorm} in CLIP~\cite{radford2021learning}, and \texttt{LayerScale} together with \texttt{RelativePositionBias} in BEiTv2~\cite{peng2022beit}.
These native components define the scaling, normalization, and positional context under which the teacher's attention was learned. 
Without them, transferring attention into the standard student may instantiate the teacher's routing patterns through a mismatched computation pathway.

\vspace{-6pt}
\paragraph{Experimental Protocol.}
To test this hypothesis, we augment the student architecture by adding only the teacher-native missing components, while keeping all other implementation details identical to~\Cref{subsec:exp_setup}.
The inserted components are initialized \textit{randomly} and trained jointly with the student, with no pre-trained weights loaded.
We evaluate this fix under both Attention Copy and Attention Distillation for all 4 representative failure teachers.
For reference, we additionally include the native-architecture No-Transfer (NoT) baseline, where the augmented student is randomly initialized and trained from scratch.

\begin{wraptable}[10]{r}{0.48\textwidth}
    \vspace{-1.2em}
    \centering
\small
\caption{\textbf{No-Transfer baseline comparison.}
We compare from-scratch No-Transfer baselines between standard student and native-arch student.
The native-arch baseline is lower than the standard-arch one, confirming the rescue is not driven by the added components' own capacity.
}
\vspace{-3pt}
\label{tab:native_not}
\ra{1.02}
\setlength{\tabcolsep}{2.5pt}
\footnotesize
\begin{tabular}{lccc}
 & \color[HTML]{d6604d}DINOv2-S & \color[HTML]{d6604d}CLIP-B & \color[HTML]{d6604d}BEiTv2-B \\
\midrule
Standard-arch & 63.1 & 69.0 & 69.0 \\
Native-arch  & 60.1 {\scriptsize\color[HTML]{d6604d}-3.0} & 66.7 {\scriptsize\color[HTML]{d6604d}-2.3} & 66.3 {\scriptsize\color[HTML]{d6604d}-2.7} \\
\end{tabular}

    \vspace{-1.5em}
\end{wraptable}

\begin{figure}[t]
    \centering
    \includegraphics[width=0.98\linewidth]{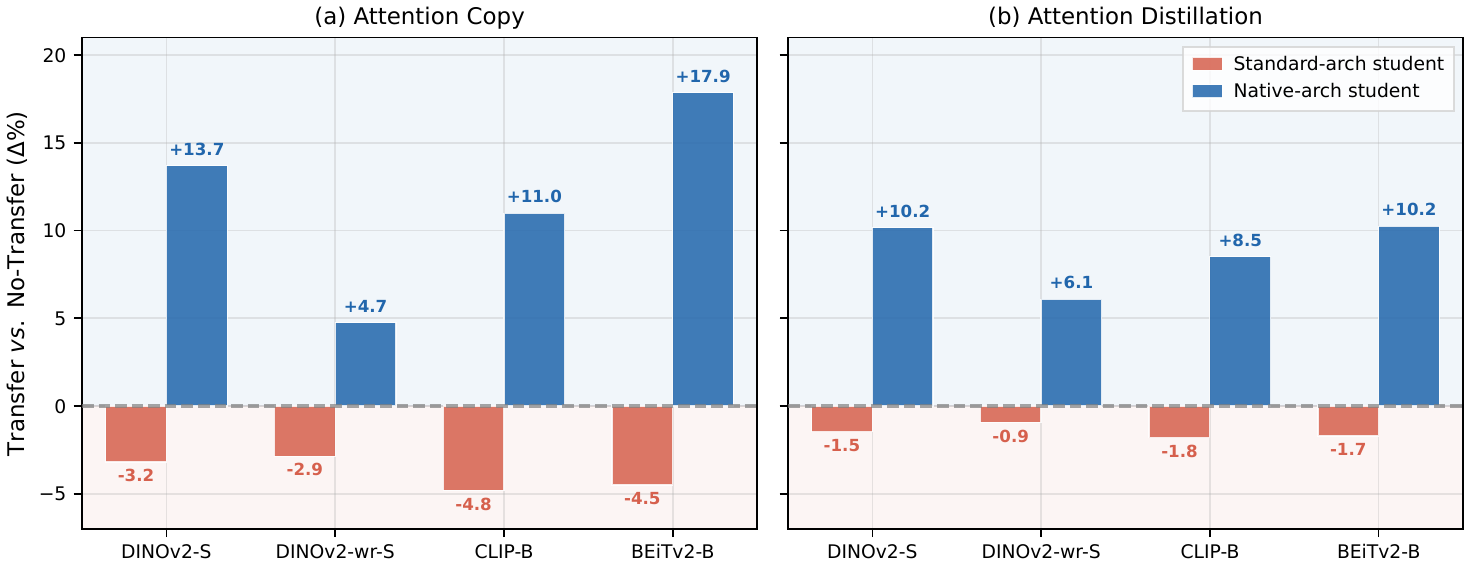}
    \vspace{-0.5em}
    \caption{\textbf{Architectural compatibility rescues ineffective Attention Transfer.}
    For teachers from the failure families, we report $\Delta$ accuracy (\%) relative to the corresponding No-Transfer baselines for both Attention Copy and Attention Distillation.
    The native-arch student is augmented by adding only the teacher-native components missing from the standard student in a randomly initialized state, with no pre-trained weights loaded.
    Across all failure families and both transfer methods, standard-arch transfer remains harmful, while native-arch transfer consistently flips to positive gains, confirming that the failure is driven by teacher-student architectural mismatch.
    }
    \label{fig:rescue}
    \vspace{-1.5em}
\end{figure}

\vspace{-6pt}
\paragraph{Results \& Analysis.}
\Cref{fig:rescue} demonstrates a clear and consistent performance flip for both Attention Transfer methods across the 4 failure families: transfer is harmful with the standard student architecture, while beneficial once teacher-native components are added.
However, one natural concern is that these gains could come from the added components' own capacity, rather than from restored compatibility with the teacher's attention patterns.
To rule this out, we compare NoT baselines for standard and native architectures.
As shown in~\Cref{tab:native_not}, native-arch NoT baselines yield worse performance than standard-arch ones for all failure families, indicating that the added components do not by themselves improve performance.
The rescue effect therefore comes specifically from restoring architectural compatibility and unlocking effective Attention Transfer.

%

\subsection{Why Architectural Mismatch Causes Failure}

\begin{figure}[t]
    \centering
    \includegraphics[width=0.98\linewidth]{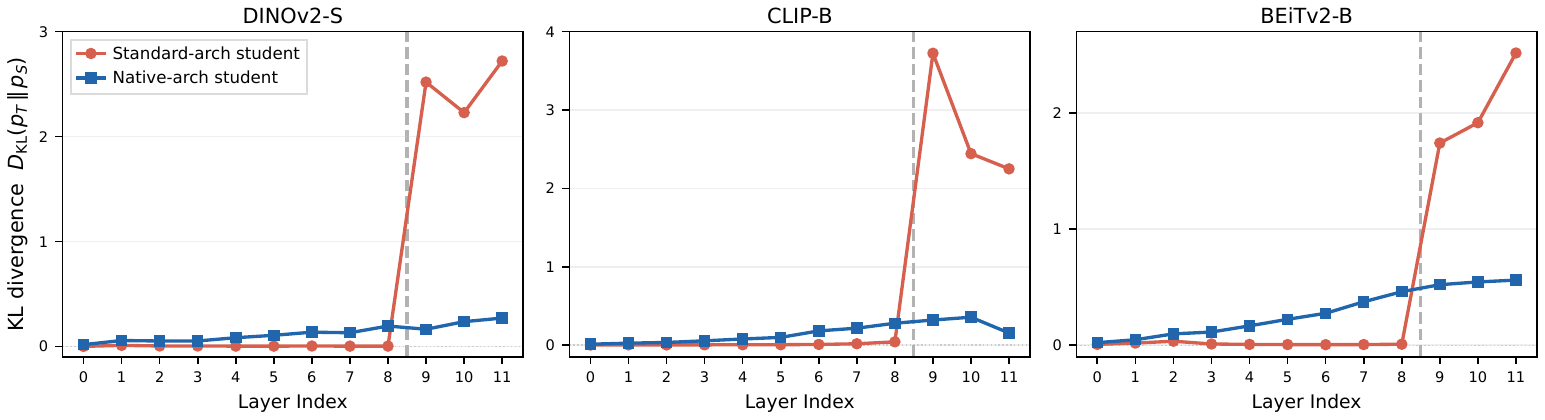}
    \vspace{-0.5em}
    \caption{\textbf{Layer-wise teacher-to-student KL divergence.}
    We plot per-layer KL divergence between the pre-trained teacher's attention maps and the transferred student's attention maps on ImageNet-1K under Attention Distillation.
    A clear spike of the standard-arch student at late layers can be observed, while the native-arch student consistently stays near-zero.
    This shows Attention Transfer can match teacher attention only when the architecture is compatible, further explaining the rescue in~\Cref{fig:rescue}.
    }
    \label{fig:student_teacher_drift}
\end{figure}

\vspace{-3pt}
To explain \emph{why} this rescue is effective, we further test the architectural compatibility hypothesis directly at the attention level: given that the teacher's pre-trained weights computed attention under its native components, a standard student lacking these components utilizes a different attention computation pathway and would be likely to fail to match the teacher's attention.
We measure this via per-layer KL divergence between the transferred student's attention and the teacher's native attention, for both standard-arch and native-arch students under Attention Distillation.
As shown in~\Cref{fig:student_teacher_drift}, the standard-arch student matches the early layers almost perfectly with near-zero KL across all three failure families, but exhibits a sharp divergence spike in the last three layers.
By contrast, the native-architecture student remains consistently low and smooth across all layers, without any collapse for transferring attention.

These results further explain the previously observed failure families: architectural mismatch in standard Attention Transfer leads to inconsistent late-layer attention learning, whereas adding native components resolves this compatibility issue and can then recover the transfer gains.
Moreover, this also clarifies the observation in~\Cref{fig:layerwise}: From-top subsets always include the unmatched final layers, whereas From-bottom subsets stay in the matchable early-layer region, explaining why From-top consistently underperforms From-bottom.
More results are given in~\Cref{subsec:appendix_arch_mismatch}.

Together, these results support architectural compatibility as the primary mechanism behind Attention Transfer's effectiveness: attention is sufficient to recover the full benefit of the teacher's pre-trained weights \textit{only} when the student architecture provides the same attention routing as the teacher's.

\section{Ruling Out Alternative Explanations}
\label{sec:alt_explanations}

\begin{figure}[t]
    \centering
    \includegraphics[width=0.7\linewidth]{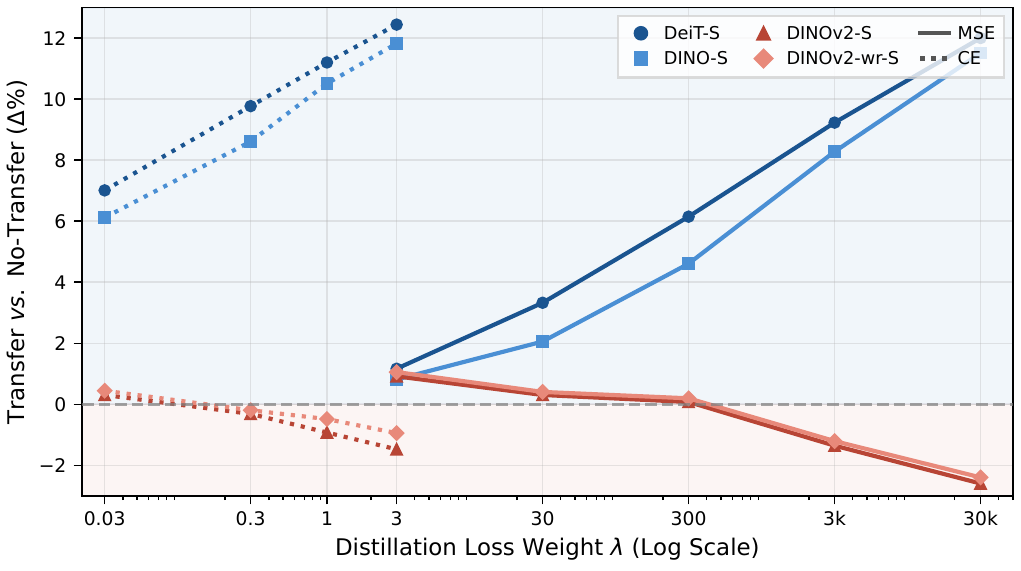}
    \vspace{-0.5em}
    \caption{\textbf{Ablation on behaviors of MSE loss \vs default CE loss.}
    We compare behaviors of different transfer losses in Attention Distillation with varying $\lambda$.
    Same family-level trends can be observed under the $\lambda$-sweep. 
    For success teachers, both losses yield monotonic gains with increasing $\lambda$; while for failure teachers, both losses cause larger drops with increasing $\lambda$.
    }
    \vspace{-1.5em}
    \label{fig:loss_sweep}
\end{figure}

\vspace{-3pt}
By establishing architectural compatibility as the primary mechanism in the previous discussion, we now ask whether simpler explanations could account for the observed failure.
We therefore investigate three natural alternative explanations below.

\vspace{-3pt}
\subsection{Not the Transfer Loss}
\label{subsec:not_the_loss}

One natural explanation for the failure is the usage of a potential unsuitable loss in Attention Transfer.
Specifically, Cross-Entropy (CE) loss is used by default for Attention Distillation, which might not be generally effective for transferring all families of pre-trained weights, and switching to a different transfer loss could rescue the failure ones.
We therefore first use MSE loss as one alternative to study the behaviors of different losses with varying $\lambda$\footnote{We follow~\cite{li2024attention} to denote $\lambda$ as the weight of the distillation loss in the overall training objective of Attention Distillation.} on representative success and failure teachers.

As shown in~\Cref{fig:loss_sweep}, MSE exhibits the same family-level trends as CE when the two losses' magnitudes are matched by adjusting $\lambda$.
For success teachers (DeiT-S, DINO-S), both losses produce monotonic improvements as $\lambda$ increases, and MSE approaches the substantial same level of performance as CE when sufficient magnitude is given by large $\lambda$.
For failure teachers (DINOv2-S, DINOv2-wr-S), both losses instead produce monotonic degradation as $\lambda$ increases.
This shows that increasing the Attention Transfer signal helps success families but amplifies harm for failure families, regardless of whether the transfer loss is CE or MSE.
We further verify this pattern across additional losses in~\Cref{subsec:appendix_transfer_loss}, all demonstrating the same pattern as MSE.
These results rule out the inadequate choice of transfer loss as the primary explanation for the observed failure.

\vspace{-3pt}
\subsection{Not the Pre-training Recipe}
\label{subsec:not_the_recipe}

Another natural explanation is that this family-dependent failure may stem from common properties of the teachers' pre-training recipes.
We test several candidate properties in the pre-training below.
Specifically, we summarize three main categories of hypotheses, including different pre-training signals, different data sources, and other known special attributes such as attention sinks and patch interpolation.
As shown in~\Cref{tab:recipe_counterexamples}, for each candidate explanation from the failure families, we can identify at least one counter-example with that property from the success families, or one teacher that lacks it but still fails.
Overall, none of these single-factor recipe properties predicts the observed family-level failure boundary.
This rules out pre-training recipe differences as the primary explanation.
Detailed discussion of these recipe properties is provided in~\Cref{subsec:appendix_recipe}, together with further analysis on controlled experiments that would fully isolate the pre-training recipe.

\begin{table}[t]
    \centering
\small
\caption{\textbf{Counter-examples for potential pre-training recipe properties.}
We list the candidate properties below from three main categories. 
For each candidate, we identify either one success teacher sharing the same property as the failure teacher, or one teacher without this property but still fails.
None of the tested single-factor recipe properties predicts the family-level failure boundary.
}
\vspace{3pt}
\label{tab:recipe_counterexamples}
\ra{1.02}
\setlength{\tabcolsep}{4.0pt}
\footnotesize
\begin{tabular}{p{0.24\linewidth}p{0.36\linewidth}p{0.36\linewidth}}
\textbf{Hypothesis} & \textbf{Failure} & \textbf{Counter-example} \\
\midrule
Pre-training signal 
& DINOv2 (Self-Distillation) & DINO (Self-Distillation)\\
& CLIP (Contrastive) & SigLIP2 (Contrastive)\\
& BEiTv2 (Masked Image Modeling) & iBOT/MAE (Masked Image Modeling)\\
\addlinespace
Data source 
& DINOv2 (Non-ImageNet Data) & SAM (Non-ImageNet Data)\\
& CLIP (Multi-Modal Data) & SigLIP2 (Multi-Modal Data)\\
\addlinespace
Special attribute
& DINOv2 (with Attention Sink) & DeiT (with Attention Sink)\\
& DINOv2 (patch $14\hspace{-2pt}\rightarrow\hspace{-2pt}16$ to fit standard-arch) & CLIP (native patch 16, also fails)\\
\end{tabular}

    \vspace{-1.8em}
\end{table}

\vspace{-3pt}
\subsection{Not the Attention Map Quality}
\label{subsec:not_the_quality}

The third natural explanation is that the failure families simply have attention maps of poorer intrinsic quality, thus a student could hardly benefit from them.
Prior evidence already argues against this: the attention of MAE is known to exhibit strong locality with weak semantic structure~\cite{xie2023revealing}, yet MAE transfers successfully, while strong semantic localization is a headline property of DINOv2~\cite{oquab2023dinov2}, yet DINOv2 fails most severely.
To test this directly, we measure the quality of attention maps with the object localization protocol of SimPool~\cite{psomas2023simpool}: MaxBoxAccV2~\cite{choe2020evaluating} on 2,000 ImageNet-1K validation images with ground-truth boxes.
More details are given in~\Cref{subsec:appendix_attn_quality}.

\begin{wraptable}[18]{r}{0.5\textwidth}
    \vspace{-1.3em}
    \centering
\small
\caption{\textbf{Attention map quality comparison before Attention Transfer.}
We report MaxBoxAccV2 (\%) of each teacher's final-layer attention under the SimPool~\cite{psomas2023simpool} localization protocol.
The quality ranking is interleaved with the success/failure split.
}
\vspace{-3pt}
\label{tab:quality_teacher}
\ra{1.02}
\setlength{\tabcolsep}{2.5pt}
\footnotesize
\begin{tabular}{l cccc}
 & \color[HTML]{2166ac}DINO-B & \color[HTML]{d6604d}DINOv2-B & \color[HTML]{d6604d}CLIP-B & \color[HTML]{2166ac}DeiT-B \\
\midrule
MaxBoxAccV2 & 79.9 & 69.8 & 68.9 & 51.6 \\
\end{tabular}
    \vspace{0.5em}
    \centering
\small
\caption{\textbf{Attention map quality comparison of transferred students w/ and w/o the rescue.}
We report MaxBoxAccV2 (\%) of the transferred students under Attention Distillation for three failure teachers.
The rescue shows no consistent effect direction on the attention map quality.
}
\vspace{-3pt}
\label{tab:quality_student}
\ra{1.02}
\setlength{\tabcolsep}{2.5pt}
\footnotesize
\begin{tabular}{l ccc}
& \color[HTML]{d6604d}DINOv2-S & \color[HTML]{d6604d}CLIP-B & \color[HTML]{d6604d}BEiTv2-B \\
\midrule
w/o rescue  & 76.1 & 72.0 & 68.2 \\
w/ rescue   & 83.7 {\scriptsize\color[HTML]{2166ac}+7.6} & 67.9 {\scriptsize\color[HTML]{d6604d}-4.1} & 51.6 {\scriptsize\color[HTML]{d6604d}-16.6} \\
\end{tabular}
    \vspace{-1.5em}
\end{wraptable}

We first evaluate the pre-trained teachers before the transfer.
As shown in~\Cref{tab:quality_teacher}, the success teacher DINO-B scores higher than both failure teachers, while the other success teacher DeiT-B scores below them: the quality ranking is interleaved with the success/failure split, and the same holds at every divergence-spike layer as shown in~\Cref{subsec:appendix_attn_quality}.
The intrinsic quality of the teacher's attention maps therefore does not predict the family-dependent failure.

Furthermore, the proposed native-arch rescue does not work through attention map quality.
Under Attention Copy, the student's attention maps are the teacher's by construction, so we compare the students under Attention Distillation, where the student learns its own attention.
As shown in~\Cref{tab:quality_student}, the rescue has no consistent effect direction on the attention map quality: it raises the measured quality under DINOv2-S but lowers it for CLIP-B and BEiTv2-B. 
In contrast, this rescue can reverse the transfer failure for all three weights as shown in~\Cref{fig:rescue}.
This cleanly separates the two explanations: the rescue works through restoring architectural compatibility, rather than through changing the attention map quality.
These results thus rule out attention map quality as the primary explanation for the observed failure.

After ruling out the inadequate choice of transfer loss, differences in pre-training recipe, and the quality of attention maps, we find that none of these alternatives accounts for the family-level failure boundary.
Taken together with~\Cref{sec:solution}, these analyses further support architectural mismatch as the primary mechanism behind the failure, further refining the prevailing understanding of attention in ViT representations: attention is sufficient \textit{only} 
when the student architecture matches the teacher.  

\vspace{-6pt}
\section{Related Work}
\label{sec:related_work}

\vspace{-3pt}
\paragraph{ViT Variants.}

Since the Transformer was introduced to vision by~\cite{dosovitskiy2020image}, a diverse landscape of pre-training paradigms has been proposed, including 
supervised training~\cite{touvron2021training,touvron2021going,touvron2022three,touvron2022deit}, 
self-distillation~\cite{caron2021emerging,oquab2023dinov2,zhou2021ibot,simeoni2025dinov3},
contrastive learning~\cite{he2020momentum,chen2020improved,chen2021empirical}, 
multi-modal pre-training~\cite{radford2021learning,jia2021scaling,zhai2023sigmoid,tschannen2025siglip}, 
masked image modeling~\cite{he2022masked, bao2022beit, peng2022beit, xie2022simmim}, 
and large-scale segmentation pre-training~\cite{kirillov2023segment,ravi2024sam,carion2025sam},
with further analyses trying to understand the underlying mechanisms behind their effectiveness~\cite{walmer2023teaching,park2023what,dravid2023rosetta,qin2025beyond}.
Several modern ViT families introduce architectural components beyond the standard design, including
LayerScale in~\cite{touvron2021going,oquab2023dinov2},
PreLayerNorm in~\cite{radford2021learning}, 
learnable relative position bias in~\cite{bao2022beit,peng2022beit},
and RoPE in~\cite{heo2024rotary,simeoni2025dinov3}.
A separate line of work has analyzed the structural role of attention itself in ViTs, aiming to mitigate attention sinks via register tokens~\cite{darcet2023vision,jiang2025testtime} or sliding windows~\cite{xiao2023streamingllm}.
However, the role of attention design in determining ViT effectiveness remains fundamentally unclear.

\vspace{-6pt}
\paragraph{Attention Transfer.}

The term \emph{Attention Transfer} was originally coined for distilling spatial activation maps in CNNs~\cite{zagoruyko2017at}, while related work in language modeling demonstrated that self-attention distributions alone can serve as a powerful distillation signal~\cite{wang2020minilm}.
This idea was later extended to the vision domain: to guide self-supervised ViT students~\cite{wang2022attention}, or to initialize ViT attention layers~\cite{trockman2023mimetic}.
Building on this line, \cite{li2024attention} further demonstrated that transferring only the model's attention patterns is sufficient to recover the full benefit of its pre-trained weights.
Some contemporary directions have explored richer feature transfer beyond attention~\cite{fan2024scalekd,trockman2026mimetic,wang2025data,tian2026low,peng2026revisiting,lin2025perspective}, initializing smaller ViTs from larger ones~\cite{xu2024initializing}, or transferring across attention operators~\cite{qin2026copy,zhang2024hedgehog,zhang2025lolcats}.
While these methods motivate Attention Transfer's appeal as a minimal-signal alternative, evaluations in~\cite{li2024attention} are limited to a few families compared to the growing diverse landscape of ViTs.

\vspace{-6pt}
\section{Conclusion}
\label{sec:conclusion}

\vspace{-3pt}
In this work, we revisit the recent claim that Attention Transfer~\cite{li2024attention} is sufficient to recover the full benefit of the pre-trained weights for Vision Transformers.
On a comprehensive model zoo of 20 teachers from 11 well-known ViT families, we reveal a hidden validity boundary: 4 families consistently fail under the standard teacher-student architecture, falling up to 5.1\% below the from-scratch no-transfer baseline, with the failure persisting across model sizes, extended training durations, different transfer datasets, and out-of-distribution evaluations.
Through a three-level localization analysis and a simple native-architecture rescue, we identify architectural mismatch between teacher and student as the primary mechanism behind this failure, and further rule out the inadequate choice of transfer loss, differences in pre-training recipe, and the quality of attention maps as alternative explanations.
Our findings refine the prevailing understanding of attention in ViT representations: 
attention is not architecture-agnostically transferable, and is sufficient \textit{only} when the student architecture preserves the same attention routing as the teacher's.
We hope our explorations, together with the corresponding analyses, could encourage the community to revisit the underlying mechanisms of Attention Transfer and further unlock the potential of this transfer paradigm.

\vspace{-6pt}

\begin{ack}
\vspace{-3pt}
This research is supported by 
the NSF of China (Grant No. 62336003, 12371510) and
the Shanghai Municipal Special Program for Basic Research on General AI Foundation Models (Grant No. 2025SHZDZX025G12),
and 
is conducted in collaboration with ST Engineering Geo-Insights Pte Ltd.
The authors sincerely thank the anonymous Reviewers and Area Chairs for their valuable feedback and constructive comments,
and declare there are no competing interests related to this research.
\end{ack}

\pagebreak

{
\small
\bibliographystyle{unsrt}
\bibliography{sections/11_references}
}

\newpage
\appendix \appendix

\cftsetindents{section}{0em}{2.1em}
\cftsetindents{subsection}{2.1em}{2.9em}
\cftsetindents{subsubsection}{5.0em}{3.8em}
\setlength{\cftbeforesecskip}{6pt}
\setlength{\cftbeforesubsecskip}{2pt}
\setlength{\cftbeforesubsubsecskip}{2pt}

\setcounter{section}{0}
\setcounter{table}{0}
\setcounter{figure}{0}
\setcounter{equation}{0}

\ifreview \nolinenumbers \fi
\begin{center}
\resizebox{\textwidth}{!}{\textbf{\paperTitle}} \\
\vspace{0.5em}\large Technical Appendices and Supplementary Material \\
\vspace{1.0em}
\end{center}
\ifreview \linenumbers \fi


\renewcommand{\thesection}{\Alph{section}}
\renewcommand{\thetable}{\Alph{table}}
\renewcommand{\thefigure}{\Alph{figure}}
\renewcommand{\theequation}{\Alph{equation}}

\tableofcontents
\newpage

\section{Additional Implementation Details}
\label{sec:appendix_implementation_details}

We provide further implementation details to support reproducibility of our experimental setup below.

\subsection{Details of Model Zoo}
\label{subsec:appendix_model_zoo}

\Cref{tab:model_zoo} summarizes all 20 pre-trained weights from 11 ViT families used in our evaluations, covering three model sizes (Tiny, Small, Base) where official checkpoints are available.
The benchmark spans the major pre-training paradigms in recent ViT representation learning, including 
supervised training~\cite{touvron2021training}, 
self-distillation~\cite{caron2021emerging,oquab2023dinov2,darcet2023vision}, 
contrastive learning~\cite{chen2021empirical}, 
multi-modal pre-training~\cite{radford2021learning,tschannen2025siglip}, 
masked image modeling~\cite{he2022masked,zhou2021ibot,peng2022beit}, 
and large-scale segmentation pre-training~\cite{kirillov2023segment}.

\subsection{Details of Experimental Setup}
\label{subsec:appendix_exp_setup_details}

\Cref{tab:training_recipe} summarizes the training recipes used for Attention Copy and Attention Distillation in our evaluations.
We follow the protocol in~\cite{li2024attention} to ensure direct comparability with their findings, with the following deliberate modifications for our broader evaluations: 
(i) we use 20-epoch training for the main benchmark and further confirm in~\Cref{fig:100ep_gap} that this is sufficient to reveal the family-level success/failure boundary;
(ii) all reported results are averaged over 3 random seeds to ensure statistical significance.
All hyper-parameters are held identical across all 20 teachers, ensuring no per-family tuning artifacts.
All experiments are conducted on NVIDIA A100 GPUs. 

\section{Additional Results and Analysis}

\subsection{Additional Analysis on Architectural Mismatch}
\label{subsec:appendix_arch_mismatch}

\begin{figure}[t]
    \centering
    \vspace{-0.5em}
    \includegraphics[width=0.98\linewidth]{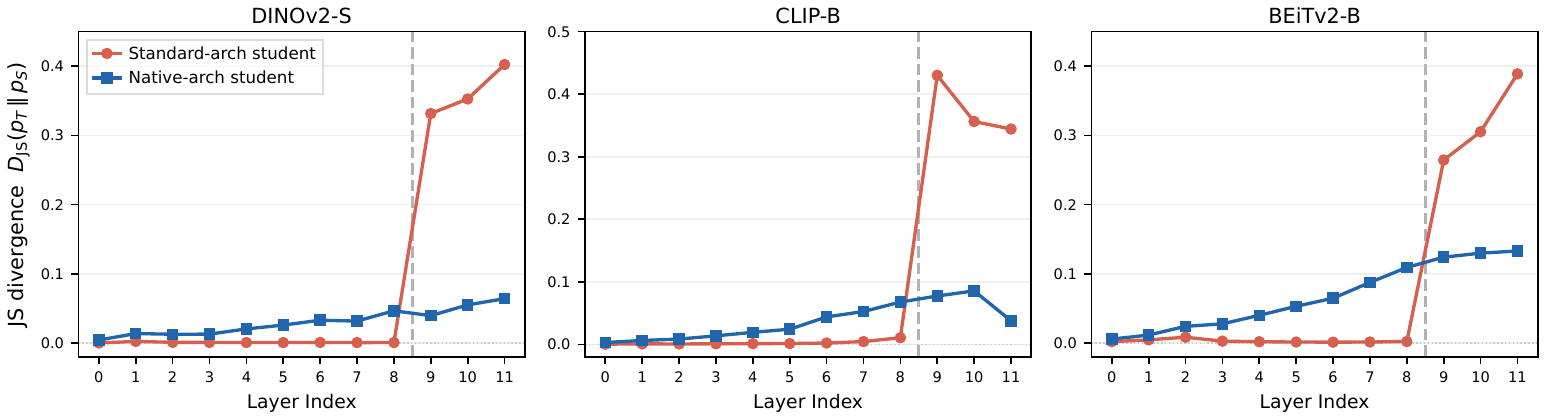}
    \vspace{-0.5em}
    \caption{\textbf{Layer-wise teacher-to-student JS divergence.}
    We plot per-layer JS divergence between the pre-trained teacher's attention maps and the transferred student's attention maps on ImageNet-1K under Attention Distillation.
    The qualitative pattern is identical to that in~\Cref{fig:student_teacher_drift}: standard-arch students show a sharp late-layer spike across all three failure families (DINOv2-S, CLIP-B, BEiTv2-B), while native-arch students stay low and smooth.
    }
    \vspace{-1.5em}
    \label{fig:student_teacher_drift_jsd}
\end{figure}

\paragraph{Layer-wise Divergence.}
The teacher-to-student divergence in~\Cref{fig:student_teacher_drift} is computed with per-layer attention maps from both the transferred student and the pre-trained teacher.
Unlike~\cite{li2024attention}, which compares different transfer methods against fine-tuned models using Jensen-Shannon (JS) divergence, our focus is to assess whether the transferred student can match the pre-trained teacher's attention for both standard-arch and native-arch students.
We use Kullback-Leibler (KL) divergence for this assessment under Attention Distillation and additionally report JS divergence in~\Cref{fig:student_teacher_drift_jsd}.
The JS curves demonstrate the same pattern as KL curves, further supporting that the architectural mismatch in standard Attention Transfer leads to inconsistent late-layer attention learning.

\begin{wraptable}[10]{r}{0.48\textwidth}
    \vspace{-1.2em}
    \centering
\small
\caption{\textbf{CLS \vs patch decomposition of the late-layer divergence.}
We report the per-layer KL and JS divergence between the teacher and standard-arch student under Attention Distillation on DINOv2-S.
}
\vspace{-3pt}
\ra{1.02}
\setlength{\tabcolsep}{2.5pt}
\footnotesize
\begin{tabular}{l cccc}
Layer & 0-8 (avg) & 9 & 10 & 11 \\
\midrule
KL on CLS   & $<$ 0.01  & 1.84 & 2.40 & 1.27 \\
KL on Patch & $<$ 0.01  & 2.52 & 2.23 & 2.73 \\
\midrule
JS on CLS   & $<$ 0.001 & 0.34 & 0.40 & 0.25 \\
JS on Patch & $<$ 0.001 & 0.33 & 0.35 & 0.40 \\
\end{tabular}
\label{tab:cls_patch}
    \vspace{-1.5em}
\end{wraptable}

\paragraph{CLS \vs Patch Decomposition.}
The divergences discussed above are computed by averaging over all 197 query tokens, which include one CLS token and 196 patch tokens.
Since the late layers are also where the CLS token becomes most important for classification, we further decompose the per-layer divergence into the CLS query row and the patch query rows for further investigation.
We test this with the representative failure teacher, DINOv2-S, with the standard-arch student under Attention Distillation.

As shown in~\Cref{tab:cls_patch}, both the CLS row and the patch rows exhibit near-zero divergence in the early layers and high divergence in the last three layers, with comparable magnitudes, and the patch rows diverging more strongly in two of the three spike layers.
The late-layer mismatch therefore exists in both the global routing and the local routing, rather than being concentrated in one type of token, which is consistent with our hypothesis: the missing native component modulates the routing of all tokens, so the induced mismatch is expected to be distributed across the entire attention computation.

Moreover, based on these decomposition results, we expect that replacing the CLS readout with global average pooling in the standard student would not fully resolve the failure, since the patch-level incompatibility inside the attention computation remains unchanged. 
Given the training cost of this diagnostic, we would like to leave the empirical validation of this expectation as future work.

\begin{table}[ht]
    \centering
\small
\caption{\textbf{Pre-trained Model Zoo.} 
We present 20 pre-trained weights from 11 well-known ViT families used in our Attention Transfer evaluations.
All weights are with the patch size of 16, while for families only with patch size of 14~\cite{oquab2023dinov2,darcet2023vision}, we interpolate their patch embeddings from 14 to 16.
}
\vspace{3pt}
\setlength{\tabcolsep}{5pt}
\ra{1.02}
\footnotesize
\begin{tabular}{l c l l c}
\textbf{Family} & \textbf{Size} & \textbf{Pre-training Data} & \textbf{Pre-training Paradigm} & \textbf{Native-Arch Components} \\
\midrule
\multicolumn{5}{l}{\textit{Success families}} \\
DeiT~\cite{touvron2021training}                & T, S, B & ImageNet-1K~\cite{deng2009imagenet}     & Supervised             & --\\
DINO~\cite{caron2021emerging}                  & S, B    & ImageNet-1K~\cite{deng2009imagenet}                                & Self-Distillation                  & -- \\
MoCov3~\cite{chen2021empirical}                & T, S, B & ImageNet-1K~\cite{deng2009imagenet}                                & Contrastive (momentum)                  & -- \\
MAE~\cite{he2022masked}                        & T, B    & ImageNet-1K~\cite{deng2009imagenet}                                & Masked Image Modeling                   & -- \\
iBOT~\cite{zhou2021ibot}                       & S, B    & ImageNet-1K~\cite{deng2009imagenet}                                & Self-Distillation (masked)                & -- \\
SigLIP2~\cite{tschannen2025siglip}             & B       & WebLI~\cite{tschannen2025siglip}         & Multi-Modal Contrastive          & -- \\
SAM~\cite{kirillov2023segment}                 & B       & SA-1B~\cite{kirillov2023segment}         & Segmentation Pre-training               & -- \\
\midrule
\multicolumn{5}{l}{\textit{Failure families}} \\
DINOv2~\cite{oquab2023dinov2}                  & S, B    & LVD-142M~\cite{oquab2023dinov2}          & Self-Distillation + MIM                 & \texttt{LayerScale} \\
DINOv2-wr~\cite{darcet2023vision}              & S, B    & LVD-142M~\cite{oquab2023dinov2}                                  & + Register Tokens                       & \texttt{LayerScale} \\
CLIP~\cite{radford2021learning}                & B       & WIT-400M~\cite{radford2021learning}      & Multi-Modal Contrastive                  & \texttt{PreLayerNorm} \\
BEiTv2~\cite{peng2022beit}                     & B       & ImageNet-1K~\cite{deng2009imagenet}                        & Masked Image Modeling                   & \texttt{LayerScale}, \texttt{RelativePositionBias} \\
\end{tabular}
\label{tab:model_zoo}
    \vspace{-1.5em}
\end{table}
\vspace{0.5em}
\begin{table}[ht]
    \centering
\small
\caption{\textbf{Training recipes for Attention Copy and Attention Distillation.}}
\vspace{3pt}
\ra{1.02}
\footnotesize
\begin{tabular}{l l l}
    \textbf{Config} & \textbf{Attention Copy} & \textbf{Attention Distillation} \\
    \midrule
    Optimizer & AdamW~\cite{loshchilov2017decoupled} & AdamW~\cite{loshchilov2017decoupled} \\
    Base Learning Rate & 1e-3 & 1e-4 \\
    Weight Decay & 0.05 & 0.3 \\
    Optimizer Momentum & $\beta_1=0.9$, $\beta_2 = 0.999$ & $\beta_1=0.9$, $\beta_2 = 0.95$ \\
    Layer-wise LR Decay & 0.75 & -- \\
    Batch Size & 1024 & 1024 \\
    LR Schedule & cosine decay & cosine decay \\
    Warm-up Epochs & 5 & 5 \\
    Epochs & 20 & 20 \\
    Augmentation & RandAug (9, 0.5) & RandAug (9, 0.5) \\
    Label Smoothing & 0.1 & 0.1 \\
    Mixup & 0.8 & 0.8 \\
    Cutmix & 1.0 & 1.0 \\
    Drop Path & 0 & 0.1 \\
    EMA & 0.9999 & 0.9999 \\
    Distillation Loss & -- & Cross-Entropy (default) \\
    Distillation Loss Weight $\lambda$ & -- & 3 \\
    Random Seeds & 3 & 3 \\
\end{tabular}
\label{tab:training_recipe}
    \vspace{-1.5em}
\end{table}

\subsection{Additional Analysis on Native-architecture Rescue}
\label{subsec:appendix_component_decomp}

\begin{wraptable}[6]{r}{0.5\textwidth}
    \vspace{-1.2em}
    \centering
\small
\caption{\textbf{Per-component decomposition of the native-arch rescue on BEiTv2-B.}
}
\vspace{-3pt}
\ra{1.02}
\setlength{\tabcolsep}{2.5pt}
\footnotesize
\begin{tabular}{l cccc}
$\Delta$ Acc & Standard & +\texttt{LayerScale} & +\texttt{RPB} & +Both \\
\midrule
Copy    & {\color[HTML]{d6604d}-4.5} & {\color[HTML]{d6604d}-2.1} & {\color[HTML]{d6604d}-1.3} & {\color[HTML]{2166ac}+17.9} \\
Distill & {\color[HTML]{d6604d}-1.7} & {\color[HTML]{d6604d}-1.3} & {\color[HTML]{d6604d}-0.8} & {\color[HTML]{2166ac}+10.2} \\
\end{tabular}
\label{tab:beitv2_decomp}
    \vspace{-1.5em}
\end{wraptable}

The native-architecture rescue in~\Cref{fig:rescue} provides results for adding all the teacher-native components missing from the standard student.
Specifically, three of the four failure families have only one single native component: the native-arch student adds only \texttt{LayerScale} for DINOv2 and DINOv2-wr, and only \texttt{PreLayerNorm} for CLIP.
The only multi-component case is BEiTv2, whose pre-trained weights natively contain both \texttt{LayerScale} and \texttt{RelativePositionBias}.
We therefore further decompose the rescue experiment for BEiTv2-B by evaluating two additional students that add only one of the two native components.
As shown in~\Cref{tab:beitv2_decomp}, each single component alone only reduces the harm but never reverses the failure, whereas adding both flips the transfer to strongly positive gains under both transfer methods.
This is consistent with our hypothesis that transferred attention remains effective only when the student preserves the full architectural context under which the teacher's attention was learned.

\subsection{Additional Analysis on Late-layer Interventions}
\label{subsec:appendix_late_layer}

Motivated by the late-layer divergence spike in~\Cref{fig:student_teacher_drift} and~\Cref{fig:student_teacher_drift_jsd}, we ask whether the failure can be fixed by changing the distillation scheme itself, rather than restoring native components as in the native-arch rescue.
We evaluate three interventions on the failure teacher DINOv2-S under the standard student:
\textit{From-bottom} transfers only the first 9 layers, leaving the last 3 layers free to learn their own attention, as in~\Cref{fig:layerwise}.
\textit{Reuse}, inspired by Skip-Attention~\cite{venkataramanan2024skip}, transfers the first 9 layers as usual while the last 3 layers reuse the teacher's layer-8 attention map, which tests whether the harmful part of the transfer is the late-layer attention content itself.
\textit{Reweight} retains supervision on all 12 layers but calibrates the distillation weight of the last 3 layers by $w$, aiming to test whether the problematic layers are merely under-supervised.

\begin{wraptable}[13]{r}{0.5\textwidth}
    \vspace{-1.2em}
    \centering
\small
\caption{\textbf{Late-layer interventions within the standard-arch on DINOv2-S.}
We report $\Delta$ accuracy (\%) relative to the corresponding No-Transfer baselines.
None of the late-layer interventions recovers the transfer benefit, while only the native-arch rescue does.
}
\vspace{-3pt}
\ra{1.02}
\setlength{\tabcolsep}{8pt}
\footnotesize
\begin{tabular}{l cc}
$\Delta$ & Copy & Distill \\
\midrule
Full transfer (default)      & {\color[HTML]{d6604d}-3.2} & {\color[HTML]{d6604d}-1.5} \\
From-bottom, 9 layers        & {\color[HTML]{d6604d}-3.7} & {\color[HTML]{d6604d}-3.2} \\
Reuse                        & {+0.8} & {\color[HTML]{d6604d}-5.4} \\
Reweight, $w=10$             & --   & {\color[HTML]{d6604d}-8.6} \\
\midrule
Native-arch rescue   & {\color[HTML]{2166ac}+13.7} & {\color[HTML]{2166ac}+10.2} \\
\end{tabular}
\label{tab:late_layer}
    \vspace{-1.5em}
\end{wraptable}

As shown in~\Cref{tab:late_layer}, none of these interventions recovers the transfer benefit.
Under Attention Copy, reusing early attention removes most of the harm but recovers none of the benefit, leaving the result at the No-Transfer level, whereas the native-arch rescue achieves a substantial improvement with the same teacher maps.
Under Attention Distillation, reusing early attention becomes even more harmful than the default full transfer, as it forces the student to learn inappropriate token routing at the wrong depth, and strengthening the supervision exactly on the problematic layers makes the failure worse still.
Together, these results indicate that avoiding the late-layer targets cannot effectively recover the transfer benefit, while only restoring the native architectural components does. 
This further supports our interpretation that the architectural mismatch is the key to the observed failure, rather than the late-layer attention content alone.

\subsection{Additional Analysis on Attention Map Quality}
\label{subsec:appendix_attn_quality}

\begin{wraptable}[10]{r}{0.48\textwidth}
    \vspace{-1.2em}
    \centering
\small
\caption{\textbf{Per-layer attention map quality comparison.}
We report per-layer MaxBoxAccV2 (\%) for the last 3 divergence-spike layers.
The success/failure ranking remains interleaved: this quality does not split the boundary.
}
\vspace{-3pt}
\ra{1.02}
\setlength{\tabcolsep}{2.5pt}
\footnotesize
\begin{tabular}{l cccc}
 & \color[HTML]{2166ac}DINO-B & \color[HTML]{d6604d}DINOv2-B & \color[HTML]{d6604d}CLIP-B & \color[HTML]{2166ac}DeiT-B \\
\midrule
Layer 9          & \textbf{84.8} & 70.1 & 58.2 & 64.1 \\
Layer 10         & 81.1 & \textbf{82.7} & 65.1 & 56.8 \\
Layer 11 & \textbf{79.9} & 69.8 & 68.9 & 51.6 \\
\end{tabular}
\label{tab:attn_quality_teacher}
    \vspace{-1.5em}
\end{wraptable}

We provide more details on the attention map quality measurement used in~\Cref{subsec:not_the_quality} and further extend it to per-layer measurement.
Following the object localization protocol of SimPool~\cite{psomas2023simpool}, we compute MaxBoxAccV2~\cite{choe2020evaluating} on the first 2,000 ImageNet-1K validation images with ground-truth boxes.
The CLS-to-patch attention map of a given layer is averaged over heads, then bilinearly upsampled to the input resolution with min-max normalization.
For each threshold $\tau\in\{0,0.01,\dots,0.99\}$, the map is binarized and every connected component is converted into a box.
An image is counted as correctly localized at IoU threshold $\delta$ if any estimated box overlaps any ground-truth box with IoU $\ge\delta$, MaxBoxAccV2 is then computed by averaging the best accuracy over $\tau$ across $\delta\in\{0.3,0.5,0.7\}$.
Teachers are evaluated with their native architectures, and students are taken from the same Attention Distillation runs as in~\Cref{fig:rescue}.

\Cref{tab:attn_quality_teacher} extends the final-layer teacher measurement of~\Cref{tab:quality_teacher} to each of the 3 divergence-spike layers of~\Cref{fig:student_teacher_drift} and~\Cref{fig:student_teacher_drift_jsd}.
As shown, at every layer, the success/failure ordering remains interleaved and no layer exhibits a quality split aligned with the transfer boundary.
The earlier spike layers also show no consistent quality degradation separating the failure families from the successes: DINOv2-B's layer 10 is its best-scoring layer, and CLIP-B's quality gradually rises toward the final layer.

\subsection{Additional Analysis on Relation to Knowledge Distillation}
\label{subsec:appendix_kd}

Our work is a mechanistic re-evaluation of the claim in~\cite{li2024attention}, instead of a new distillation method: 
\cite{li2024attention} claims that transferring only the attention patterns can recover the full benefit of the pre-trained weights, and our contribution is to establish \emph{when} this claim holds and \emph{why} it fails.
Standard knowledge distillation methods are therefore \textit{not} baselines that our conclusion competes against.
The two lines of work also operate in different paradigms: 
standard knowledge distillation typically serves model compression, where a task-specific teacher transfers its logits or features to a smaller student, whereas Attention Transfer operates in the pre-training paradigm, transferring only the attention from a pre-trained teacher into a matched-size, randomly initialized student in order to isolate which part of the pre-trained knowledge suffices.
A direct comparison is further complicated in our benchmark by the fact that most teachers are self-supervised or vision-language pre-trained models without a classifier, so logit distillation is not directly defined unless each teacher is first fine-tuned, which would mainly alter the object of study.
We also note that the comparison between Attention Transfer and feature distillation is provided in~\cite{li2024attention}: distilling the features of the pre-trained teacher recovers only a small fraction of the benefit recovered by Attention Distillation within the matched-size scope.

\begin{wraptable}[9]{r}{0.5\textwidth}
    \vspace{-1.2em}
    \centering
\small
\caption{\textbf{Attention Distillation \vs feature distillation comparison.}
We report $\Delta$ accuracy (\%) relative to the No-Transfer baselines of the corresponding students, following the feature distillation implementation in~\cite{li2024attention}.
}
\vspace{-3pt}
\ra{1.02}
\setlength{\tabcolsep}{2.1pt}
\footnotesize
\begin{tabular}{l ccc}
$\Delta$ & \color[HTML]{2166ac}MAE-B & \color[HTML]{d6604d}DINOv2-B & \color[HTML]{2166ac}DINOv2-B (w/ rescue) \\
\midrule
Attn. Distill   & \color[HTML]{2166ac}{+8.1} & \color[HTML]{d6604d}{-2.3} & \color[HTML]{2166ac}{+9.5} \\
Feat. Distill   & \color[HTML]{2166ac}{+2.2} & \color[HTML]{2166ac}{+1.7} & \color[HTML]{2166ac}{+5.1} \\
\end{tabular}
\label{tab:kd}
    \vspace{-1.5em}
\end{wraptable}

Moreover, we extend this comparison under our protocol on one success teacher and one failure teacher, together with the native-arch rescue.
As shown in~\Cref{tab:kd}, for the success family, feature distillation recovers only a small fraction of the Attention Transfer gain, reproducing the pattern in~\cite{li2024attention}.
For the failure family, Attention Distillation harms the standard student while feature distillation still transfers positively; however, the native-architecture rescue flips Attention Distillation to outperform feature distillation again.
In other words, whether Attention Transfer outperforms standard distillation is itself governed by architectural compatibility, which is precisely the validity boundary our paper establishes.

\subsection{Additional Analysis on Transfer Loss}
\label{subsec:appendix_transfer_loss}

\begin{figure}[ht]
    \centering
    \includegraphics[width=0.7\linewidth]{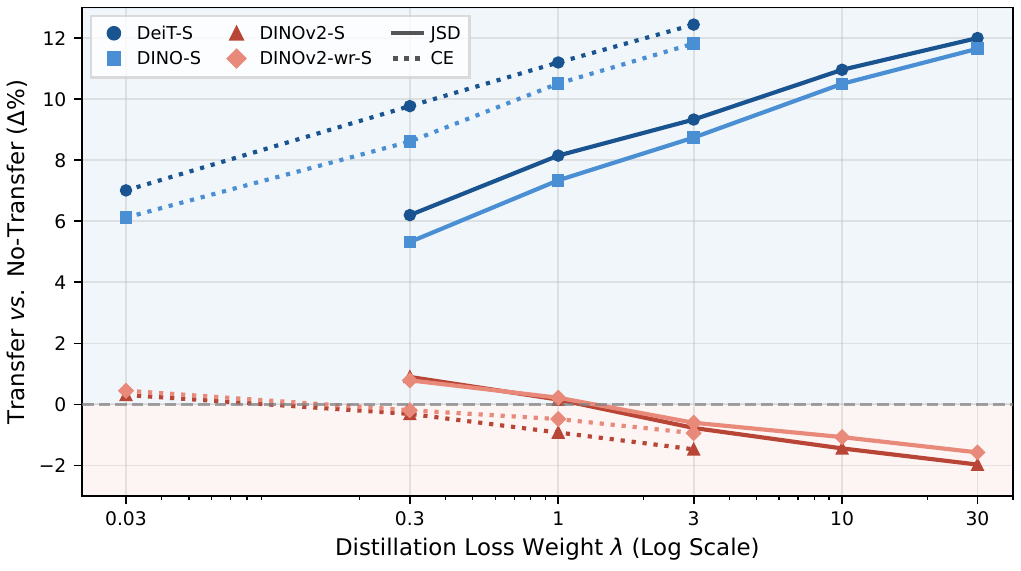}
    \vspace{0.4em}
    \includegraphics[width=0.7\linewidth]{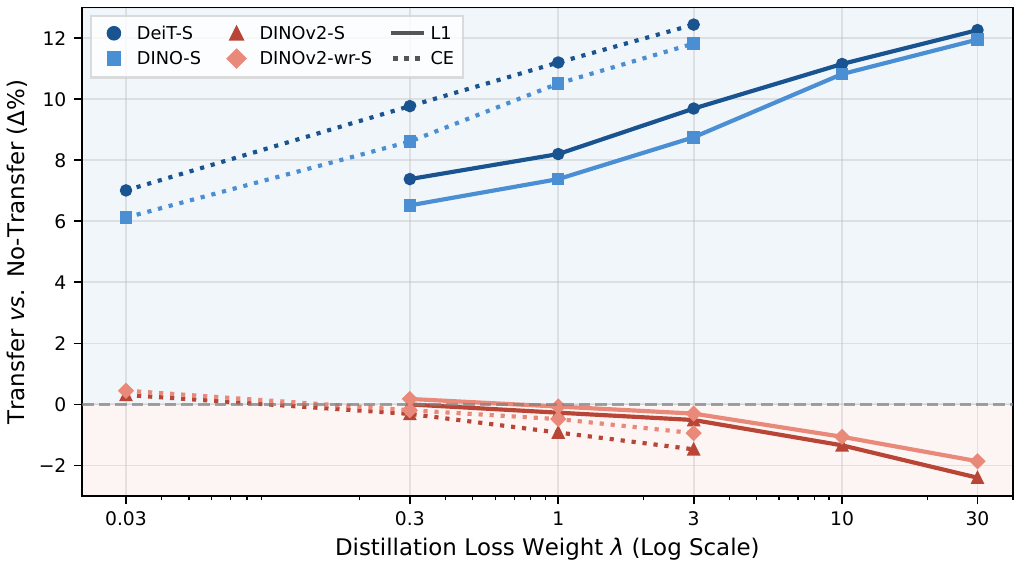}
    \vspace{-0.5em}
    \caption{\textbf{Ablation on behaviors of JSD loss (\textit{top}) / L1 loss (\textit{bottom}) \vs default CE loss.}
    We compare behaviors of different transfer losses in Attention Distillation with varying $\lambda$.
    Same family-level trends can be observed under the $\lambda$-sweep. 
    For success teachers, both losses yield monotonic gains with increasing $\lambda$; while for failure teachers, both losses cause larger drops with increasing $\lambda$.
    }
    \label{fig:loss_sweep_jsd_l1}
    \vspace{-1.5em}
\end{figure}

In addition to the CE loss \vs MSE loss discussed in~\Cref{subsec:not_the_loss}, we further verify whether the observed family-level failure persists under other well-known distillation losses with distinct geometric properties.
We extend the $\lambda$-sweep on the same 4 representative teachers using two additional losses: the Jensen-Shannon divergence, and the L1 distance.
Plots in~\Cref{fig:loss_sweep_jsd_l1} together show that the family-level success/failure split observed in~\Cref{fig:loss_sweep} remains consistent under both JSD and L1.
For success teachers (DeiT-S, DINO-S), all four losses produce monotonic gains with increasing $\lambda$; for failure teachers (DINOv2-S, DINOv2-wr-S), all produce monotonic degradation.
Notably, both the success-side gain and the failure-side harm scale monotonically with the loss's effective gradient magnitude, but the sign of the family-level split is invariant.
This further confirms that the family-level failure is not an artifact of the inadequate choice of transfer loss, complementing the discussion in~\Cref{subsec:not_the_loss}.

\subsection{Additional Analysis on Pre-training Recipe}
\label{subsec:appendix_recipe}

We provide more detailed discussion on the candidate pre-training-recipe hypotheses summarized in~\Cref{tab:recipe_counterexamples}.
For each candidate, we identify either (i) one success teacher that shares the same property with one failure teacher, or (ii) one teacher without the property that still fails.

\paragraph{Pre-training Signal.}
One natural hypothesis is that the failure families share a particular pre-training objective that the standard student cannot effectively transfer.
We find that the 4 failure families cover the following three categories:
\begin{itemize}
    \item Self-Distillation: DINOv2 and DINOv2-wr both fail under Attention Transfer, while DINO, which also uses a self-distillation paradigm based on an EMA teacher, succeeds.

    \item Contrastive Multi-Modal Pre-training: CLIP fails under Attention Transfer despite being pre-trained with multi-modal contrastive learning, while SigLIP2, which also uses the same pre-training paradigm, succeeds.

    \item Masked Image Modeling: BEiTv2 fails under masked image modeling, while iBOT and MAE, which both perform masked-image modeling during pre-training, succeed.
\end{itemize}

The pre-training signal therefore does not predict the family-level failure boundary.

\paragraph{Data Source.}
The second hypothesis is that the failure families share a particular group of training data, because most ViT weights are trained on ImageNet-1K, with the analysis below:
\begin{itemize}
    \item Non-ImageNet Data: DINOv2 from failure families is pre-trained on LVD-142M, a curated non-ImageNet dataset of $\sim$142M images.
    However, SAM is also pre-trained on one non-ImageNet dataset, SA-1B, yet succeeds.
    
    \item Multi-Modal Data:  CLIP is pre-trained on a multi-modal image-text dataset, WIT-400M, and fails, while SigLIP2 is also pre-trained on a multi-modal dataset, WebLI, yet succeeds.
\end{itemize}

The pre-training data source therefore does not predict the family-level failure boundary.

\paragraph{Special Attribute.}
The third hypothesis is that the failure families share special structural properties in their attention or specific implementation tricks.
\begin{itemize}
    \item Attention Sink: DINOv2 has been found to develop strong attention sinks during pre-training~\cite{darcet2023vision}, which has been speculated to disrupt downstream transferability of attention. 
    However, attention sinks are also well-documented in standard supervised ViTs such as DeiT~\cite{darcet2023vision,xiao2023streamingllm}, yet DeiT succeeds.
    
    \item Patch-Size Interpolation: DINOv2 is officially released with patch size 14, which we adapt to patch size 16 via patch-embedding interpolation to fit the standard teacher-student architecture protocol. 
    One may suspect that this interpolation introduces representational artifacts that cause the observed failure. 
    However, CLIP is natively released with patch size 16 and is used with no interpolation in our setup, yet still fails under Attention Transfer.
\end{itemize}

Taken together, none of these candidate single-factor properties of pre-training recipes individually predicts the previously observed family-level failure boundary.
This further supports our central claim in~\Cref{sec:solution} that the family-level boundary is driven by architectural mismatch between the pre-trained teacher and the standard student, rather than by any specific aspect of the pre-training recipe.

\paragraph{Towards Fully Controlled Experiments.}
The above analysis is based on counter-examples rather than fully controlled pre-training, where such an experiment would isolate the architectural factor from the pre-training recipe at the teacher side, complementing our native-arch rescue, which isolates it at the student side.
For a failure family, the controlled experiment would fix the pre-training data, objective, and all hyper-parameters, and pre-train two teachers that differ \emph{only} in the native architectural components:
one with the original native architecture, and one with the standard ViT architecture.
Our hypothesis is that Attention Transfer from the standard-arch teacher into the standard student would then behave like the success families.
For a success family, the controlled experiment would add the native components to the teacher under the same pre-training recipe, with the hypothesis that Attention Transfer from this modified teacher into the standard student would then degrade toward the failure pattern.
Since the original pre-training data of failure families such as DINOv2 and CLIP is not publicly available, such experiments would have to be conducted on open-source data at a reduced scale.
We believe this is a valuable direction and leave it for future work.

\subsection{Additional Analysis on Sensitivity of Attention Distillation Failure}
\label{subsec:appendix_sensitivity}

We further clarify the robustness of the observed failure from the following aspects:
First, to ensure a fair comparison across all 20 teachers, we strictly keep all training hyper-parameters identical by following the default settings of~\cite{li2024attention}, which were originally tuned to make Attention Transfer succeed; no per-family tuning is involved.
Second, the family-level failure is equally reflected by Attention Copy, which involves no distillation-specific optimization at all. 
The same failure pattern under both Attention Transfer methods indicates that the failure is not created by distillation-specific optimization choices.
Third, the transfer objective in Attention Distillation is jointly optimized with the downstream task training.
Changing the learning rate, schedule, or warm-up also changes the entire optimization, including the No-Transfer baseline itself, which could also affect the effectiveness of general task optimization.
For a controlled sensitivity analysis, the adjustable transfer-specific factors are therefore only the distillation loss weight and the choice of loss, both of which are swept extensively in~\Cref{subsec:not_the_loss} and~\Cref{subsec:appendix_transfer_loss} to show that the family-level failure never changes.
Finally, one remaining accessible robustness test is longer training, which we have examined in~\Cref{fig:100ep_gap}: the failure remains consistent throughout and tends to converge by 100 epochs, confirming that the negative transfer is not an artifact of insufficient optimization.

\subsection{Additional Analysis on Statistical Significance}
\label{subsec:appendix_statistical_significance}

We further provide the statistical significance analysis of our main results in~\Cref{fig:main_results}.
We report the per-teacher standard deviation of $\Delta$ accuracy under both Attention Copy and Attention Distillation for all 6 pre-trained weights with model size Small across 3 random seeds in~\Cref{tab:main_results_std}.

\begin{table}[t]
    \centering
\small
\caption{\textbf{Statistical significance analysis on main results.}
We report the per-teacher standard deviation of our main results on all 6 pre-trained weights with model size Small over 3 random seeds.
The standard deviations are uniformly within 0.2\%, far below the magnitude of every reported $\Delta$, confirming that the observed family-level failure boundary in~\Cref{fig:main_results} is robust to seed variation. 
}
\vspace{3pt}
\setlength{\tabcolsep}{4pt}
\ra{1.02}
\footnotesize
\begin{tabular}{c c c c c c c}
& \textcolor[HTML]{2166ac}{DeiT-S}
& \textcolor[HTML]{2166ac}{DINO-S}
& \textcolor[HTML]{2166ac}{MoCov3-S}
& \textcolor[HTML]{2166ac}{iBOT-S}
& \textcolor[HTML]{d6604d}{DINOv2-S}
& \textcolor[HTML]{d6604d}{DINOv2-wr-S} \\
\midrule
$\Delta$ Acc (Copy)        & +12.2 & +11.1 & +4.8 & +11.9 & -3.2 & -2.9 \\
$\pm$ std           & 0.13   & 0.17   & 0.10  & 0.15   & 0.07  & 0.11  \\
\midrule
$\Delta$ Acc (Distill)    & +12.4 & +12.0 & +7.5 & +12.1 & -1.5 & -0.9 \\
$\pm$ std       & 0.14   & 0.16   & 0.11  & 0.13   & 0.09  & 0.10  \\
\label{tab:main_results_std}
\end{tabular}

    \vspace{-2.5em}
\end{table}

\section{Limitations \& Scope}
\label{sec:appendix_limitations}

Our findings are scoped to Attention Transfer protocol studied in~\cite{li2024attention}: transferring softmax attention patterns from a pre-trained ViT teacher to a randomly initialized ViT student through either Attention Copy or Attention Distillation.
Within this setting, our results show that the effectiveness of Attention Transfer is governed by architectural compatibility between teacher and student.
This claim concerns only softmax-to-softmax attention-pattern transfer, rather than all possible ways of reusing pre-trained ViT weights or converting attention mechanisms.
Other transfer paradigms, such as full-weight initialization, feature/knowledge distillation, parameter-efficient fine-tuning, or transferring softmax attention to linear or sparse attention variants, fall outside the scope of our conclusions.

The architectural compatibility principle concerns whether the student ViT preserves the architectural context under which the teacher learned its attention patterns.
For the failure families identified in our benchmark, this context is captured by concrete teacher-native components, including \texttt{LayerScale}, \texttt{PreLayerNorm}, and \texttt{RelativePositionBias}, which are introduced by different ViT variants.
Our native-architecture rescue shows that restoring these components in randomly initialized form is sufficient to make the transferred attention usable, so we interpret them as architectural scaffolding rather than additional pre-trained knowledge.
Our experiments use matched-size teacher–student pairs, and the findings should therefore be read as characterizing architectural compatibility under matched-capacity transfer rather than cross-size transfer.

Our evaluations focus on classification-centered Attention Transfer settings, strictly following the protocol of~\cite{li2024attention}, including ImageNet-1K, iNaturalist, and ImageNet-based out-of-distribution benchmarks.
These settings test whether the observed failure boundary persists across different training durations, transfer datasets, and distribution shifts.
Applications where attention plays a different operational role, such as dense prediction, vision-language transfer, or cross-resolution deployment, may require task-specific validation before applying the same conclusion.
Our mechanistic evidence is empirical: component-level decomposition, native-architecture rescue, native-scratch controls, and teacher-student attention divergence consistently support architectural compatibility as the primary mechanism behind the observed failure.
A complete formal theory of Attention Transfer remains an important direction for future work.

\section{Broader Impacts}
\label{sec:appendix_broader_impact}

This work is primarily a methodological study of when Attention Transfer between pre-trained ViTs succeeds or fails.
Its direct positive impact is to encourage more careful reporting of teacher-student architectural assumptions in transfer and distillation studies, and to help the community avoid ineffective transfer attempts under mismatched architectures.
A potential negative impact is misinterpretation: our findings should not be read as ``Attention Transfer does not work'', but rather that Attention Transfer is effective only when the student architecture is compatible with the teacher.
Finally, this paper relies on existing public pre-trained weights, some of which are trained on large-scale curated or web-scale datasets and may inherit dataset biases.
Architectural compatibility can make transfer more effective, but it does not remove biases or safety concerns inherited from the original teacher weights.

\newpage
\section*{NeurIPS Paper Checklist}

\begin{enumerate}

\item {\bf Claims}
    \item[] Question: Do the main claims made in the abstract and introduction accurately reflect the paper's contributions and scope?
    \item[] Answer: \answerYes{} 
    \item[] Justification: The main claims in the abstract and introduction accurately reflect the paper's contribution and scope. We revisit the recent claim that Attention Transfer is sufficient to recover the full benefit of the pre-trained weights of Vision Transformers and reveal that Attention Transfer is not universally effective. Further experiments confirm that our findings refine the prevailing understanding of attention in ViT representations: attention is sufficient only when the student architecture matches the teacher.
    \item[] Guidelines:
    \begin{itemize}
        \item The answer \answerNA{} means that the abstract and introduction do not include the claims made in the paper.
        \item The abstract and/or introduction should clearly state the claims made, including the contributions made in the paper and important assumptions and limitations. A \answerNo{} or \answerNA{} answer to this question will not be perceived well by the reviewers. 
        \item The claims made should match theoretical and experimental results, and reflect how much the results can be expected to generalize to other settings. 
        \item It is fine to include aspirational goals as motivation as long as it is clear that these goals are not attained by the paper. 
    \end{itemize}

\item {\bf Limitations}
    \item[] Question: Does the paper discuss the limitations of the work performed by the authors?
    \item[] Answer: \answerYes{} 
    \item[] Justification: We have discussed the limitations of our work in~\Cref{sec:appendix_limitations}.
    \item[] Guidelines:
    \begin{itemize}
        \item The answer \answerNA{} means that the paper has no limitation while the answer \answerNo{} means that the paper has limitations, but those are not discussed in the paper. 
        \item The authors are encouraged to create a separate ``Limitations'' section in their paper.
        \item The paper should point out any strong assumptions and how robust the results are to violations of these assumptions (e.g., independence assumptions, noiseless settings, model well-specification, asymptotic approximations only holding locally). The authors should reflect on how these assumptions might be violated in practice and what the implications would be.
        \item The authors should reflect on the scope of the claims made, e.g., if the approach was only tested on a few datasets or with a few runs. In general, empirical results often depend on implicit assumptions, which should be articulated.
        \item The authors should reflect on the factors that influence the performance of the approach. For example, a facial recognition algorithm may perform poorly when image resolution is low or images are taken in low lighting. Or a speech-to-text system might not be used reliably to provide closed captions for online lectures because it fails to handle technical jargon.
        \item The authors should discuss the computational efficiency of the proposed algorithms and how they scale with dataset size.
        \item If applicable, the authors should discuss possible limitations of their approach to address problems of privacy and fairness.
        \item While the authors might fear that complete honesty about limitations might be used by reviewers as grounds for rejection, a worse outcome might be that reviewers discover limitations that aren't acknowledged in the paper. The authors should use their best judgment and recognize that individual actions in favor of transparency play an important role in developing norms that preserve the integrity of the community. Reviewers will be specifically instructed to not penalize honesty concerning limitations.
    \end{itemize}

\item {\bf Theory assumptions and proofs}
    \item[] Question: For each theoretical result, does the paper provide the full set of assumptions and a complete (and correct) proof?
    \item[] Answer: \answerNA{} 
    \item[] Justification: We did not include any theoretical results in this work.
    \item[] Guidelines:
    \begin{itemize}
        \item The answer \answerNA{} means that the paper does not include theoretical results. 
        \item All the theorems, formulas, and proofs in the paper should be numbered and cross-referenced.
        \item All assumptions should be clearly stated or referenced in the statement of any theorems.
        \item The proofs can either appear in the main paper or the supplemental material, but if they appear in the supplemental material, the authors are encouraged to provide a short proof sketch to provide intuition. 
        \item Inversely, any informal proof provided in the core of the paper should be complemented by formal proofs provided in appendix or supplemental material.
        \item Theorems and Lemmas that the proof relies upon should be properly referenced. 
    \end{itemize}

    \item {\bf Experimental result reproducibility}
    \item[] Question: Does the paper fully disclose all the information needed to reproduce the main experimental results of the paper to the extent that it affects the main claims and/or conclusions of the paper (regardless of whether the code and data are provided or not)?
    \item[] Answer: \answerYes{} 
    \item[] Justification: We have included the experimental setups in~\Cref{subsec:exp_setup} and~\Cref{sec:appendix_implementation_details}.
    \item[] Guidelines:
    \begin{itemize}
        \item The answer \answerNA{} means that the paper does not include experiments.
        \item If the paper includes experiments, a \answerNo{} answer to this question will not be perceived well by the reviewers: Making the paper reproducible is important, regardless of whether the code and data are provided or not.
        \item If the contribution is a dataset and\slash or model, the authors should describe the steps taken to make their results reproducible or verifiable. 
        \item Depending on the contribution, reproducibility can be accomplished in various ways. For example, if the contribution is a novel architecture, describing the architecture fully might suffice, or if the contribution is a specific model and empirical evaluation, it may be necessary to either make it possible for others to replicate the model with the same dataset, or provide access to the model. In general. releasing code and data is often one good way to accomplish this, but reproducibility can also be provided via detailed instructions for how to replicate the results, access to a hosted model (e.g., in the case of a large language model), releasing of a model checkpoint, or other means that are appropriate to the research performed.
        \item While NeurIPS does not require releasing code, the conference does require all submissions to provide some reasonable avenue for reproducibility, which may depend on the nature of the contribution. For example
        \begin{enumerate}
            \item If the contribution is primarily a new algorithm, the paper should make it clear how to reproduce that algorithm.
            \item If the contribution is primarily a new model architecture, the paper should describe the architecture clearly and fully.
            \item If the contribution is a new model (e.g., a large language model), then there should either be a way to access this model for reproducing the results or a way to reproduce the model (e.g., with an open-source dataset or instructions for how to construct the dataset).
            \item We recognize that reproducibility may be tricky in some cases, in which case authors are welcome to describe the particular way they provide for reproducibility. In the case of closed-source models, it may be that access to the model is limited in some way (e.g., to registered users), but it should be possible for other researchers to have some path to reproducing or verifying the results.
        \end{enumerate}
    \end{itemize}

\item {\bf Open access to data and code}
    \item[] Question: Does the paper provide open access to the data and code, with sufficient instructions to faithfully reproduce the main experimental results, as described in supplemental material?
    \item[] Answer: \answerNo{} 
    \item[] Justification: To ensure reproducibility, we will open-source the code upon publication.
    \item[] Guidelines:
    \begin{itemize}
        \item The answer \answerNA{} means that paper does not include experiments requiring code.
        \item Please see the NeurIPS code and data submission guidelines (\url{https://neurips.cc/public/guides/CodeSubmissionPolicy}) for more details.
        \item While we encourage the release of code and data, we understand that this might not be possible, so \answerNo{} is an acceptable answer. Papers cannot be rejected simply for not including code, unless this is central to the contribution (e.g., for a new open-source benchmark).
        \item The instructions should contain the exact command and environment needed to run to reproduce the results. See the NeurIPS code and data submission guidelines (\url{https://neurips.cc/public/guides/CodeSubmissionPolicy}) for more details.
        \item The authors should provide instructions on data access and preparation, including how to access the raw data, preprocessed data, intermediate data, and generated data, etc.
        \item The authors should provide scripts to reproduce all experimental results for the new proposed method and baselines. If only a subset of experiments are reproducible, they should state which ones are omitted from the script and why.
        \item At submission time, to preserve anonymity, the authors should release anonymized versions (if applicable).
        \item Providing as much information as possible in supplemental material (appended to the paper) is recommended, but including URLs to data and code is permitted.
    \end{itemize}

\item {\bf Experimental setting/details}
    \item[] Question: Does the paper specify all the training and test details (e.g., data splits, hyperparameters, how they were chosen, type of optimizer) necessary to understand the results?
    \item[] Answer: \answerYes{} 
    \item[] Justification: We have included the implementation details in~\Cref{subsec:exp_setup} and~\Cref{sec:appendix_implementation_details}.
    \item[] Guidelines:
    \begin{itemize}
        \item The answer \answerNA{} means that the paper does not include experiments.
        \item The experimental setting should be presented in the core of the paper to a level of detail that is necessary to appreciate the results and make sense of them.
        \item The full details can be provided either with the code, in appendix, or as supplemental material.
    \end{itemize}

\item {\bf Experiment statistical significance}
    \item[] Question: Does the paper report error bars suitably and correctly defined or other appropriate information about the statistical significance of the experiments?
    \item[] Answer: \answerYes{} 
    \item[] Justification: We have included the statistical significance analysis of our main results in~\Cref{subsec:appendix_statistical_significance}.
    \item[] Guidelines:
    \begin{itemize}
        \item The answer \answerNA{} means that the paper does not include experiments.
        \item The authors should answer \answerYes{} if the results are accompanied by error bars, confidence intervals, or statistical significance tests, at least for the experiments that support the main claims of the paper.
        \item The factors of variability that the error bars are capturing should be clearly stated (for example, train/test split, initialization, random drawing of some parameter, or overall run with given experimental conditions).
        \item The method for calculating the error bars should be explained (closed form formula, call to a library function, bootstrap, etc.)
        \item The assumptions made should be given (e.g., Normally distributed errors).
        \item It should be clear whether the error bar is the standard deviation or the standard error of the mean.
        \item It is OK to report 1-sigma error bars, but one should state it. The authors should preferably report a 2-sigma error bar than state that they have a 96\% CI, if the hypothesis of Normality of errors is not verified.
        \item For asymmetric distributions, the authors should be careful not to show in tables or figures symmetric error bars that would yield results that are out of range (e.g., negative error rates).
        \item If error bars are reported in tables or plots, the authors should explain in the text how they were calculated and reference the corresponding figures or tables in the text.
    \end{itemize}

\item {\bf Experiments compute resources}
    \item[] Question: For each experiment, does the paper provide sufficient information on the computer resources (type of compute workers, memory, time of execution) needed to reproduce the experiments?
    \item[] Answer: \answerYes{} 
    \item[] Justification: We have included sufficient information on the computing resources needed to reproduce the experiments in~\Cref{subsec:appendix_exp_setup_details}.
    \item[] Guidelines:
    \begin{itemize}
        \item The answer \answerNA{} means that the paper does not include experiments.
        \item The paper should indicate the type of compute workers CPU or GPU, internal cluster, or cloud provider, including relevant memory and storage.
        \item The paper should provide the amount of compute required for each of the individual experimental runs as well as estimate the total compute. 
        \item The paper should disclose whether the full research project required more compute than the experiments reported in the paper (e.g., preliminary or failed experiments that didn't make it into the paper). 
    \end{itemize}
    
\item {\bf Code of ethics}
    \item[] Question: Does the research conducted in the paper conform, in every respect, with the NeurIPS Code of Ethics \url{https://neurips.cc/public/EthicsGuidelines}?
    \item[] Answer: \answerYes{} 
    \item[] Justification: Our research adheres to all ethical guidelines required by NeurIPS.
    \item[] Guidelines:
    \begin{itemize}
        \item The answer \answerNA{} means that the authors have not reviewed the NeurIPS Code of Ethics.
        \item If the authors answer \answerNo, they should explain the special circumstances that require a deviation from the Code of Ethics.
        \item The authors should make sure to preserve anonymity (e.g., if there is a special consideration due to laws or regulations in their jurisdiction).
    \end{itemize}

\item {\bf Broader impacts}
    \item[] Question: Does the paper discuss both potential positive societal impacts and negative societal impacts of the work performed?
    \item[] Answer: \answerYes{} 
    \item[] Justification: We have discussed the broader impacts of our work in~\Cref{sec:appendix_broader_impact}.
    \item[] Guidelines:
    \begin{itemize}
        \item The answer \answerNA{} means that there is no societal impact of the work performed.
        \item If the authors answer \answerNA{} or \answerNo, they should explain why their work has no societal impact or why the paper does not address societal impact.
        \item Examples of negative societal impacts include potential malicious or unintended uses (e.g., disinformation, generating fake profiles, surveillance), fairness considerations (e.g., deployment of technologies that could make decisions that unfairly impact specific groups), privacy considerations, and security considerations.
        \item The conference expects that many papers will be foundational research and not tied to particular applications, let alone deployments. However, if there is a direct path to any negative applications, the authors should point it out. For example, it is legitimate to point out that an improvement in the quality of generative models could be used to generate Deepfakes for disinformation. On the other hand, it is not needed to point out that a generic algorithm for optimizing neural networks could enable people to train models that generate Deepfakes faster.
        \item The authors should consider possible harms that could arise when the technology is being used as intended and functioning correctly, harms that could arise when the technology is being used as intended but gives incorrect results, and harms following from (intentional or unintentional) misuse of the technology.
        \item If there are negative societal impacts, the authors could also discuss possible mitigation strategies (e.g., gated release of models, providing defenses in addition to attacks, mechanisms for monitoring misuse, mechanisms to monitor how a system learns from feedback over time, improving the efficiency and accessibility of ML).
    \end{itemize}
    
\item {\bf Safeguards}
    \item[] Question: Does the paper describe safeguards that have been put in place for responsible release of data or models that have a high risk for misuse (e.g., pre-trained language models, image generators, or scraped datasets)?
    \item[] Answer: \answerNA{} 
    \item[] Justification: This paper poses no such risks to the best of our knowledge.
    \item[] Guidelines:
    \begin{itemize}
        \item The answer \answerNA{} means that the paper poses no such risks.
        \item Released models that have a high risk for misuse or dual-use should be released with necessary safeguards to allow for controlled use of the model, for example by requiring that users adhere to usage guidelines or restrictions to access the model or implementing safety filters. 
        \item Datasets that have been scraped from the Internet could pose safety risks. The authors should describe how they avoided releasing unsafe images.
        \item We recognize that providing effective safeguards is challenging, and many papers do not require this, but we encourage authors to take this into account and make a best faith effort.
    \end{itemize}

\item {\bf Licenses for existing assets}
    \item[] Question: Are the creators or original owners of assets (e.g., code, data, models), used in the paper, properly credited and are the license and terms of use explicitly mentioned and properly respected?
    \item[] Answer: \answerYes{} 
    \item[] Justification: All datasets and pre-trained weights used in our work are publicly available, and we have cited the respective literature for each of them. Any researcher can download these datasets and pre-trained weights from the provided sources.
    \item[] Guidelines:
    \begin{itemize}
        \item The answer \answerNA{} means that the paper does not use existing assets.
        \item The authors should cite the original paper that produced the code package or dataset.
        \item The authors should state which version of the asset is used and, if possible, include a URL.
        \item The name of the license (e.g., CC-BY 4.0) should be included for each asset.
        \item For scraped data from a particular source (e.g., website), the copyright and terms of service of that source should be provided.
        \item If assets are released, the license, copyright information, and terms of use in the package should be provided. For popular datasets, \url{paperswithcode.com/datasets} has curated licenses for some datasets. Their licensing guide can help determine the license of a dataset.
        \item For existing datasets that are re-packaged, both the original license and the license of the derived asset (if it has changed) should be provided.
        \item If this information is not available online, the authors are encouraged to reach out to the asset's creators.
    \end{itemize}

\item {\bf New assets}
    \item[] Question: Are new assets introduced in the paper well documented and is the documentation provided alongside the assets?
    \item[] Answer: \answerNA{} 
    \item[] Justification: We do not release new assets. We will make the code public after the paper is accepted.
    \item[] Guidelines:
    \begin{itemize}
        \item The answer \answerNA{} means that the paper does not release new assets.
        \item Researchers should communicate the details of the dataset\slash code\slash model as part of their submissions via structured templates. This includes details about training, license, limitations, etc. 
        \item The paper should discuss whether and how consent was obtained from people whose asset is used.
        \item At submission time, remember to anonymize your assets (if applicable). You can either create an anonymized URL or include an anonymized zip file.
    \end{itemize}

\item {\bf Crowdsourcing and research with human subjects}
    \item[] Question: For crowdsourcing experiments and research with human subjects, does the paper include the full text of instructions given to participants and screenshots, if applicable, as well as details about compensation (if any)? 
    \item[] Answer: \answerNA{} 
    \item[] Justification: The paper does not involve crowdsourcing nor research with human subjects.
    \item[] Guidelines:
    \begin{itemize}
        \item The answer \answerNA{} means that the paper does not involve crowdsourcing nor research with human subjects.
        \item Including this information in the supplemental material is fine, but if the main contribution of the paper involves human subjects, then as much detail as possible should be included in the main paper. 
        \item According to the NeurIPS Code of Ethics, workers involved in data collection, curation, or other labor should be paid at least the minimum wage in the country of the data collector. 
    \end{itemize}

\item {\bf Institutional review board (IRB) approvals or equivalent for research with human subjects}
    \item[] Question: Does the paper describe potential risks incurred by study participants, whether such risks were disclosed to the subjects, and whether Institutional Review Board (IRB) approvals (or an equivalent approval/review based on the requirements of your country or institution) were obtained?
    \item[] Answer: \answerNA{} 
    \item[] Justification: The paper does not involve crowdsourcing nor research with human subjects.
    \item[] Guidelines:
    \begin{itemize}
        \item The answer \answerNA{} means that the paper does not involve crowdsourcing nor research with human subjects.
        \item Depending on the country in which research is conducted, IRB approval (or equivalent) may be required for any human subjects research. If you obtained IRB approval, you should clearly state this in the paper. 
        \item We recognize that the procedures for this may vary significantly between institutions and locations, and we expect authors to adhere to the NeurIPS Code of Ethics and the guidelines for their institution. 
        \item For initial submissions, do not include any information that would break anonymity (if applicable), such as the institution conducting the review.
    \end{itemize}

\item {\bf Declaration of LLM usage}
    \item[] Question: Does the paper describe the usage of LLMs if it is an important, original, or non-standard component of the core methods in this research? Note that if the LLM is used only for writing, editing, or formatting purposes and does \emph{not} impact the core methodology, scientific rigor, or originality of the research, declaration is not required.
    \item[] Answer: \answerNA{} 
    \item[] Justification: The core method development in this research does not involve LLMs as any important, original, or non-standard components.
    \item[] Guidelines:
    \begin{itemize}
        \item The answer \answerNA{} means that the core method development in this research does not involve LLMs as any important, original, or non-standard components.
        \item Please refer to our LLM policy in the NeurIPS handbook for what should or should not be described.
    \end{itemize}

\end{enumerate}

\end{document}